%% file: draft_main.tex
\documentclass{article}

\input{custom_command.tex}

\icmltitlerunning{Minimal Embeddable Dimensions for Top-k retrieval.}

\begin{document}

\twocolumn[
  \icmltitle{$\real^{2k}$ is Theoretically Large Enough for
  Embedding-based Top-$k$ Retrieval}



  \icmlsetsymbol{equal}{*}

  \begin{icmlauthorlist}
    \icmlauthor{Zihao Wang}{tsy}
    \icmlauthor{Hang Yin}{sqp}
    \icmlauthor{Lihui Liu}{wsu}
    \icmlauthor{Hanghang Tong}{uiuc}
    \icmlauthor{Yangqiu Song}{hkust}
    \icmlauthor{Ginny Wong}{nv}
    \icmlauthor{Simon See}{nv}
  \end{icmlauthorlist}

  \icmlaffiliation{tsy}{TSY Capital}
  \icmlaffiliation{sqp}{Squarepoint Capital}
  \icmlaffiliation{wsu}{Wayne State University, MI, USA}
  \icmlaffiliation{uiuc}{University of Illinois Urbana–Champaign, IL, USA}
  \icmlaffiliation{hkust}{Hong Kong University of Science and Technology}
  \icmlaffiliation{nv}{NVIDIA AI Technology Center (NVATIC), NVIDIA,
  Santa Clara, USA}

  \icmlcorrespondingauthor{Zihao Wang}{zihaow0710@gmail.com}

  \icmlkeywords{Machine Learning, ICML}

  \vskip 0.3in
]



\printAffiliationsAndNotice{}  

\begin{abstract}
This paper studies the Minimal Embeddable Dimension (MED): the least dimension in which there exists a configuration of $m$ object vectors so that every subset of size at most $k$ is exactly retrieved by score comparison. Our result shows MED is $\Theta(k)$, independent of $m$, for inner product, Euclidean distance, and cosine similarity. We then consider Robust MED (RMED), where all vectors are unit normed and an $\epsilon$ gap of scores is required. We derive the $m$-dependent feasibility ceiling $\epsilon_\star(m,k)=m/\sqrt{k(m-1)(m-k)}$, which approaches $1/\sqrt{k}$ when $m\gg k$, and a Gaussian centroid construction gives a robust witness upper bound in the feasible margin regime. Numerical simulation on synthetic top-$2$ retrieval with cyclic polytope and centroid query optimization confirmed our theoretical claims. Experiments on LIMIT and LIMIT-small datasets also show that simple embedding-based retrieval baselines can overfit and outperform the reported single-vector LLM embedding baseline. Both theoretical and empirical findings rule out the lack of exact geometric capacity as the obstruction.
\end{abstract}

\section{Introduction}

Embedding-based retrieval systems answer queries by vector comparison. A system stores each object $x_i\in X$ as a vector $\vx_i\in \real^d$, embeds a query $q$ as a vector $\vw_q\in \real^d$, scores each object by $s(\vx_i,\vw_q)$, and returns the desired answer set by thresholding scores, or equivalently by top-$|S|$ retrieval when the answer-set size is known. This paper studies the following approximability question behind this workflow:
\begin{tcolorbox}[colback=gray!6, colframe=black!60, sharp corners, boxrule=0.5pt]
Given a universe $X$ with \underline{$m$ objects} and a \underline{scoring function $s$}, what is the minimal \underline{dimension $d$} of $\real^d$ such that every query with \underline{at most $k$ answers} is perfectly retrievable?
\end{tcolorbox}
We call this dimension the Minimal Embeddable Dimension (MED). Informally, a configuration is embeddable if the object and query vectors can, via score comparison, realize every desired answer set. Formally, Section~\ref{sec:problem-definition} defines this as $k$-shattering by the functional class induced by the scoring function.

The question is closely related to VC dimension in statistical learning theory~\citep{mohri2018foundations}, the $k$-set problem in combinatorial geometry~\citep{matousek2013lectures}, and dimensionality limits in information retrieval, compression, and representation learning, and other topics~\citep{alon1985geometrical,lee2019latent,weller2025mfollowir,andoni2008near,izacard2021unsupervised,wang2022text}. Empirically, retrieval quality often depends on the embedding dimension when the universe size $m$ is large~\citep{yin2018dimensionality,reimers2021curse}. Recent work has argued that the dimension of vector space can constrain retrieval under positive-margin or learned-embedding protocols~\citep{weller2025theoretical} through theoretical analysis and empirical benchmarks.

We convey a significantly different message from~\citep{weller2025theoretical}: The geometric approximability is not the obstruction. The real hardness lies in optimization protocols, generalization of neural models, etc. We show this by making three points:

\noindent\textbf{Exact MED is $\Theta(k)$.}
For inner product, Euclidean distance, and cosine similarity, we prove matching-order exact MED bounds: the lower bound is $k-1$, while the upper bounds are $2k$ for inner product and Euclidean distance and $2k+1$ for cosine similarity. The lower bound comes from the VC dimension. The upper bound for the case of inner product scores comes from a construction: placing the $m$ objects on the vertices of a cyclic polytope, constructing the query vectors by a square polynomial. And the Euclidean and cosine bounds follow by reductions.

\noindent\textbf{Robust MED is $O(k^2\log m)$ in the feasible regime.}
To study the stronger requirement beyond exact approximability, we define RMED under a unit-ball normalization and require every selected object to beat every unselected object by at least $\epsilon$. In this normalized setting, positive margin changes the dimension regime: $m$ reappears through packing lower bounds~\citep{weller2025theoretical} with logarithmic growth. Feasibility is capped by $\epsilon_\star(m,k)=m/\sqrt{k(m-1)(m-k)}$; our RMED asymptotics use the retrieval regime $m/k\to\infty$, where $\epsilon_\star\sim1/\sqrt{k}$, while finite-$m$ statements retain $\epsilon_\star$. At the feasible margin scale, a Gaussian centroid construction gives an $O(k^2\log m)$ centroid-query witness at margin $c/\sqrt{k}$ for inner product and cosine, with a constant-factor Euclidean transfer.

\noindent\textbf{``Hard'' empirical benchmarks are actually easy.}
Previous empirical evidence in~\citep{weller2025theoretical} comes from two benchmarks: (1) \textit{free embedding optimization} of object and query vectors in the top-2 query test set. Their result suggested the MED grows with $m$ polynomially. The cyclic-polytope construction gives an exact dimension-$4$ top-$2$ witness for arbitrary $m$, while centroid optimization grows slowly on the tested grid. (2) \textit{LIMIT and LIMIT-small datasets} that are very easy to build but ``extremely hard'' for the state-of-the-art embedding-based retrieval model.
The reported best of single vector embedding models achieved about 0.03 and 0.53 top-2 recall scores on LIMIT and LIMIT-small in $\real^{4096}$. We use random additive constructions to sample object and query vectors without training, and surpass the ``best single vector model. '' with as few as 512 dimensions. Interestingly, we found that we can use a cyclic polytope to ``overfit'' LIMIT and LIMIT-small in $\real^4$, and if we push the random additive construction to $R^{4096}$, we achieve 0.95 and 0.70 top-2 recall on LIMIT-small and LIMIT datasets respectively.

The resulting message is not that practical retrieval is solved by a low-dimensional construction. The exact cyclic-polytope witness may have tiny margins, poor numerical conditioning, and an infeasible number of subset-specific query maps. Rather, the theory rules out a lack of exact geometric capacity as the obstruction and then identifies the stronger requirements under which geometry and learning become difficult.

\section{Minimal Embeddable Dimension}\label{sec:problem-definition}

For convenience, we summarize the notation used in this paper.
$m, n, d, k$ are positive integers, and $n$ and $d$ both denote ambient dimension. $X$ is the set of $m$ elements to be embedded and $\vx \in \real^d$ denotes the embedding of $x\in X$. For simplicity, we do not distinguish an element from its embedding when the meaning is clear, and write $X = \{\vx_i\}_{i=1}^m \subseteq \real^d$.
We study queries whose answer sets contain at least one and at most $k$ objects:
\[
    \collection_k=\{S\subseteq X: 1\leq |S|\leq k\}.
\]
The empty answer set is omitted because centroid and normalized robust queries are undefined for $|S|=0$; for exact threshold retrieval, adding the empty set would not change the results below. We also use $q$ to denote a subset in $X$.
The scoring function $s: \real^d \times \real^d\mapsto \real$ measures the relatedness of two vectors in $\real^d$. The description of $\real^d$ is omitted if the context is clear. The scoring functions of our interest include:
\begin{compactdesc}
    \item[Linear:] $s_{\rm linear}(\vx, \vw) = \langle \vx, \vw\rangle$, where $\vw$ is a query vector, $\langle \cdot, \cdot\rangle$ is the inner product.
    \item[Cosine similarity:] $s_{\cos}(\vx, \vw) = \frac{\langle \vx, \vw \rangle }{ \|\vx\|\|\vw\|}$.
    \item[Euclidean distance ($\ell_2$):] $s_{\ell_2}(\vx, \vw) = - \|\vx - \vw\|_2 $.
\end{compactdesc}
For convenience, we consider the functional classes $\functionals$ induced by those three scoring functions, e.g., the linear functional class $\fln = \{f(\cdot ):= s_{\rm linear}(\cdot, \vw) | \vw \in \real^d\}$, the cosine family $\fcos = \{  f(\cdot ):= s_{\cos}(\cdot, \vw) | \vw \in \real^d\}$, and the $\ell_2$ family $\fl{2} = \{f(\cdot ):= s_{\ell_2}(\cdot, \vw) | \vw \in \real^d\}$. Each functional $f\in \functionals$ maps $\real^d$ to $\real$. We use the subscript to indicate that the specific function $f_q\in \functionals$ is used for a specific query $q$, or $f_S$ for a subset $S$.

The primary focus of this section is to define the Minimal Embeddable Dimension (MED). We first relate exact MED to VC dimension, then define a normalized positive-margin variant, RMED. Finally, we introduce centroid-query embedding, where each subset query uses the centroid $\vc_S = |S|^{-1}\sum_{x\in S}\vx$ as its query vector. This restricted protocol gives concrete witnesses for both exact MED and robust MED.

\subsection{$k$-shatter problem}
To formally define the minimal embeddable dimension, we introduce the concept of $k$-shattering.
\begin{definition}[$k$-shattering]
    Let $X\subseteq \real^d$ be a set of $m$ points. The set $X$ is $k$-shattered by $\functionals$ if and only if, for every $S\in \collection_k$, there exist $f_S\in \functionals$ and $b_S\in \real$ such that
    \[
        f_S(\vx)>b_S>f_S(\vy)
        \qquad
        \forall \vx\in S,\ \forall \vy\in X\setminus S .
    \]
\end{definition}

\begin{remark}
    The definition of $k$-shattering precisely determines whether there exists a configuration of vectors in $X$ under a specific functional family or scoring function with query embeddings such that the embedding-based retrieval built upon this configuration succeeds on all queries concerning at most $k$ elements.
\end{remark}

For example, let $X=\{x_1,x_2,x_3,x_4\}$ and suppose a query for $S=\{x_1,x_3\}$ assigns scores
\begin{align*}
    f_S(x_1)&=0.8, & f_S(x_2)&=0.1,\\
    f_S(x_3)&=0.7, & f_S(x_4)&=-0.2 .
\end{align*}
The threshold $b_S=0.5$ exactly retrieves $S$. If the answer-set size is known, this is the same as returning the top two objects. A positive-margin version asks for more: in the normalized RMED definition below, the selected scores must exceed all unselected scores by at least a prescribed gap $\epsilon$.

Minimal Embeddable Dimension ($\med$) is then defined based on $k$-shattering, which depends on $m$, $k$, and $\functionals$. For convenience, $\med$ is denoted as a function $n^*=\med(m, k; \functionals)$.
\begin{definition}[Minimal Embeddable Dimension]\label{def:med}
    Given $m$, $k$, and $\functionals$, $\med(m,k;\functionals)$ is the smallest integer $n^*$ such that some configuration of $m$ points in $\real^{n^*}$ can be $k$-shattered by $\functionals$. If no finite dimension works, we write $\med(m,k;\functionals)=\infty$.
\end{definition}

One direct result, according to Definition~\ref{def:med}, is the non-strict monotonicity of $\med(m, k; \functionals)$.
\begin{proposition}\label{prop:monotonic}
For $2\leq k\leq m$, the following inequality holds:
\begin{align*}
    \med(m, k-1;\functionals)
    &\leq \med(m, k; \functionals) \\
    &\leq \med(m+1, k; \functionals).
\end{align*}
\end{proposition}

We don't need to discuss $k > m/2$ situations, because separating any $k$ points from $m-k$ is equivalent to separating $m-k$ points from $k$, which is already discussed in $m-k \leq m/2$ cases.

\subsection{General bounds of $\med$ by VC dimension}

The definition of $k$-shattering induces a binary threshold class
\[
    \mathcal{C}_{\functionals,n}
    =
    \bigl\{\{\vx:f(\vx)>b\}: f\in\functionals,\ b\in\real\bigr\}
\]
on $\real^n$. VC dimension~\citep{mohri2018foundations} is always taken with respect to this thresholded class.
\begin{definition}[VC dimension]
    The VC dimension $\vcd(n;\functionals)$ is the largest size of a finite set in $\real^n$ shattered by $\mathcal{C}_{\functionals,n}$.
\end{definition}
For notational convenience, define the generalized inverse
\[
    \vcd^{-1}(m;\functionals)=\min \{i : \vcd(i;\functionals)\ge m\}.
\]
The function $\vcd(\cdot ;\functionals)$ need not increase strictly, so this is not an ordinary inverse.
\begin{lemma}\label{lemma:connection-VCD}
    $\med(m, m;\functionals) = \vcd^{-1}(m;\functionals)$.
\end{lemma}

\begin{proof}
    Let $n = \vcd^{-1}(m; \functionals)$, then (1) $m$ points in $\real^n$ can be $m$-shattered by $\functionals$ but (2) $m$ points in $\real^{(n-1)}$ cannot be $m$-shattered by $\functionals$. Those two claims imply, in the sense of \med, that (1) $\med(m, m; \functionals) \leq n$ and (2) $\med(m, m; \functionals) > n - 1$. Thus $\med(m, m; \functionals) = n$.
\end{proof}

Additionally, combining Proposition~\ref{prop:monotonic} and Lemma~\ref{lemma:connection-VCD} yields the following proposition.
\begin{proposition}\label{prop:vc-bounds}
    $$\vcd^{-1}(k; \functionals) \leq \med(m, k; \functionals) \leq \vcd^{-1}(m; \functionals).$$
\end{proposition}
Thus, a rough lower and upper bound of $\med$ can be derived from the VC dimension. Specifically, the upper (lower) bounds of VC dimensions now form the lower (upper) bounds of $\med$, respectively.

\subsection{Robust MED}\label{sec:rmed-definition}
Exact $k$-shattering only requires a strict score ordering. RMED adds the first stronger requirement missing from exact geometric approximability: a normalized positive score gap. Such a gap is meaningful only after normalization; otherwise, inner-product scores can be rescaled by multiplying the query vector. We therefore define RMED under a unit-ball normalization and keep the scoring family explicit.

Let $\mathbb{B}_d=\{\vz\in\real^d:\|\vz\|_2\leq 1\}$. For $\functionals\in\{\fln,\fcos,\fl{2}\}$ induced by a score $s$, robust retrieval uses object vectors and query vectors in $\mathbb{B}_d$; for cosine similarity, the vectors are required to be nonzero, and may equivalently be normalized to the unit sphere.

\begin{definition}[$\epsilon$-robust $k$-shattering]\label{def:robust-shattering}
    Let $\functionals$ be induced by a scoring function $s$. A set $X=\{\vv_i\}_{i=1}^m\subseteq \mathbb{B}_d$ is $\epsilon$-robustly $k$-shattered by $\functionals$ if and only if for every $S\in \collection_k$ with $X\setminus S\neq\emptyset$, there exists a valid query vector $\vw_S\in \mathbb{B}_d$ such that
    \begin{align*}
        \min_{i\in S}s(\vv_i,\vw_S)
        \geq
        \max_{j\notin S}s(\vv_j,\vw_S)+\epsilon.
    \end{align*}
\end{definition}

\begin{definition}[Robust Minimal Embeddable Dimension]\label{def:rmed}
    Given $m$, $k$, $\epsilon>0$, and a functional family $\functionals$, $\rmed(m,k,\epsilon;\functionals)$ is the smallest integer $d$ such that a set of $m$ points in $\mathbb{B}_d$ can be $\epsilon$-robustly $k$-shattered by $\functionals$. If no finite dimension works, we write $\rmed(m,k,\epsilon;\functionals)=\infty$.
\end{definition}

\begin{proposition}[One-way relation between MED and RMED]\label{prop:med-rmed-one-way}
    For every $\epsilon>0$ and functional family $\functionals$, whenever $\rmed(m,k,\epsilon;\functionals)$ is finite,
    \[
        \med(m,k;\functionals)\leq \rmed(m,k,\epsilon;\functionals).
    \]
\end{proposition}
\begin{proof}
    Any $\epsilon$-robust witness has a positive selected-unselected score gap, and therefore gives a strict separating threshold for the same subset. Thus every robust realization is also an exact realization for the same score family in the same dimension.
\end{proof}
This implication is only one-way. VC-dimension lower bounds for exact MED therefore transfer to RMED as zero-margin necessary conditions, but VC dimension alone does not characterize RMED because it does not control normalized margins.

\subsection{Centroid query embedding}\label{sec:maed}
Centroid-query embedding restricts the query vector for a subset $S$ to be determined by the selected object embeddings:
\[
    \vc_S=\frac{1}{|S|}\sum_{x\in S}\vx .
\]
This is not a new threshold class; it is a construction protocol inside the scoring models above. If these fixed centroid queries retrieve every $S\in\collection_k$, then the same object configuration is a witness for exact MED. If their normalized directions retrieve every $S$ with margin $\epsilon$, then the same configuration is a witness for RMED.

\begin{definition}[Centroid-query MED]\label{def:medc}
    Given $m$, $k$, and a scoring function $s$, $\medc(m,k;s)$ is the smallest integer $d$ such that there exist $m$ points $X\subseteq\real^d$ for which, for every $S\in\collection_k$ and every $\vx\in S$, $\vy\in X\setminus S$,
    \[
        s(\vx,\vc_S)>s(\vy,\vc_S),
        \qquad
        \vc_S=\frac{1}{|S|}\sum_{x\in S}\vx .
    \]
    If no finite dimension works, we write $\medc(m,k;s)=\infty$.
\end{definition}

\begin{definition}[Robust centroid-query MED]\label{def:rmedc}
    Given $m$, $k$, $\epsilon>0$, and a functional family $\functionals$ induced by a score $s$, $\rmedc(m,k,\epsilon;\functionals)$ is the smallest integer $d$ such that there exist points $X=\{\vv_i\}_{i=1}^m\subseteq\mathbb{B}_d$ for which, for every $S\in\collection_k$ with $X\setminus S\neq\emptyset$, the centroid $\vc_S=|S|^{-1}\sum_{i\in S}\vv_i$ is nonzero and its normalized direction $\vu_S=\vc_S/\|\vc_S\|_2$ satisfies
    \[
        \min_{i\in S}s(\vv_i,\vu_S)
        \geq
        \max_{j\notin S}s(\vv_j,\vu_S)+\epsilon.
    \]
    If no finite dimension works, we write $\rmedc(m,k,\epsilon;\functionals)=\infty$.
\end{definition}

In MED and RMED, each subset query may choose its own functional or query direction. In the centroid-query variants, all query vectors are determined by the $m$ object vectors. These restrictions are useful for constructions and experiments, but they can only increase the required dimension.
\begin{proposition}[Centroid queries witness exact and robust MED]\label{prop:centroid-witnesses}
    Let $\functionals_s$ be the functional family induced by a scoring function $s$. Whenever the right-hand side is finite,
    \[
        \med(m,k;\functionals_s)\leq \medc(m,k;s).
    \]
    Whenever $\rmedc(m,k,\epsilon;\functionals_s)$ is finite,
    \[
        \rmed(m,k,\epsilon;\functionals_s)
        \leq
        \rmedc(m,k,\epsilon;\functionals_s).
    \]
    For the centroid witnesses used below, dropping the positive margin also gives the exact centroid-query relation
    \[
        \medc(m,k;s)
        \leq
        \rmedc(m,k,\epsilon;\functionals_s).
    \]
\end{proposition}
\begin{proof}
    A centroid-query MED witness supplies, for each $S$, the functional $f_S(\cdot)=s(\cdot,\vc_S)$, and therefore satisfies the exact $k$-shattering definition. A robust centroid-query witness is an $\epsilon$-robust $k$-shattering witness with the additional restriction $\vu_S=\vc_S/\|\vc_S\|_2$. Finally, the Gaussian centroid witnesses used below have unit object vectors. For linear and cosine scores, replacing the normalized centroid direction by the raw centroid only applies a positive rescaling or the same cosine direction. For Euclidean scores on unit objects, distance ranking against a centroid direction is equivalent to inner-product ranking against that direction. Hence dropping the margin gives strict exact centroid retrieval for the same object configuration.
\end{proof}

Relation to learned set embeddings. Centroid queries are a deliberately
simple aggregation protocol. More expressive query encoders, such as
neural set encoders~\citep{zaheer2017deep}, could map
$\{\vx_i:i\in S\}$ to a query vector by an MLP, attention block, or
other learned set function. Such an intermediate MED-N-style regime is
more flexible than fixed centroids but less free than arbitrary
per-subset functionals. We do not formalize it here; $\medc$ and
$\rmedc$ are used as clean construction protocols that witness MED and
RMED, not as a claim that centroids are the final practical retrieval
architecture.
\section{Exact MED: a Construction}\label{sec:optimal-standard-theory}

This section derives $\Theta(k)$ bounds for $\med$ under $\fln$, $\fcos$, and $\fl{2}$.
The safest core result is the inner-product construction: cyclic-polytopal neighborliness gives an explicit $2k$-dimensional witness for every $m$. The Euclidean and cosine statements are reductions from this inner-product result, with the cosine reduction using one extra dimension. Lower bounds come from VC dimension via Proposition~\ref{prop:vc-bounds}.

\subsection{Inner product-score bounds}\label{sec:linear-bounds}
Let us consider the cyclic polytope~\citep{ziegler2012lectures}.
\begin{example}[Cyclic polytope]\label{eg:cyclic-polytope}
	Given the moment curve $\vx(t) = (t, t^2, \dots, t^d)\in \real^d, 0\leq t\leq 1$, a cyclic polytope is the convex hull ${\rm Conv}(\vx(t_1), \vx(t_2), \dots, \vx(t_m))$, where $t_1 < t_2 < \cdots < t_m$ can be arbitrary real numbers.
\end{example}
One of the most well-known results of cyclic polytopes is that a cyclic polytope in $\real^d$ is an $\lfloor d/2 \rfloor$-neighborly polytope, meaning that every subset of at most $\lfloor d/2 \rfloor$ vertices is the vertex set of a face~\citep{ziegler2012lectures}. Equivalently, those selected vertices can be strictly separated from the remaining vertices by an affine hyperplane.

The following lemma gives an explicit construction of the separating hyperplane---and hence the query vector---for any $k$-subset of points on the moment curve. It is the coefficient form of the standard polynomial proof of cyclic-polytope neighborliness~\citep{ziegler2012lectures,matousek2013lectures}.

\begin{lemma}[Squared-polynomial query construction]\label{lem:poly-construction}
Let $t_1 < t_2 < \dots < t_m$ be distinct real numbers. Embed each element $i$ as $\vv_i = (t_i, t_i^2, \dots, t_i^n) \in \real^n$. For any subset $S \subseteq \{1,\dots,m\}$ with $|S| \le k$, define the monic polynomial
\[
P_S(t) = \prod_{i \in S} (t - t_i) = \sum_{j=0}^{|S|} a_j t^j \qquad (a_{|S|} = 1),
\]
and expand its square via coefficient convolution as $P_S^2(t) = \sum_{j=0}^{2|S|} c_j t^j$. Construct the query vector
\[
\vq_S = (-c_1, -c_2, \dots, -c_n) \in \real^n,
\]
where $c_j = 0$ for $j > 2|S|$. Then, whenever $n \ge 2k$,
\[
\langle \vv_i, \vq_S \rangle = c_0 \quad \forall i \in S,
\qquad
\langle \vv_j, \vq_S \rangle < c_0 \quad \forall j \notin S.
\]
\end{lemma}
\begin{proof}
For any $t \in \real$, $\langle (t, t^2, \dots, t^n), \vq_S \rangle = -\sum_{j=1}^{n} c_j t^j$. Since $n \ge 2k \ge 2|S|$, all non-zero coefficients of $P_S^2$ are captured, hence
\[
\langle (t, \dots, t^n), \vq_S \rangle = -(P_S^2(t) - c_0) = c_0 - P_S^2(t).
\]
For $i \in S$, $t_i$ is a root of $P_S$, so $P_S^2(t_i) = 0$ and the score equals $c_0$. For $j \notin S$, $P_S(t_j) \neq 0$, thus $P_S^2(t_j) > 0$ and the score is strictly less than $c_0$. If $X\setminus S$ is nonempty, choose $b_S$ between $\max_{j \notin S} (c_0 - P_S^2(t_j))$ and $c_0$; if $S=X$, the separation condition is vacuous on the unselected side. This yields the strict separation required by $k$-shattering.
\end{proof}

Intuitively, the squared polynomial $P_S^2(t)$ vanishes exactly on $S$ and is positive elsewhere; the query $\vq_S$ extracts its coefficients so that the inner product $\langle \vv_i, \vq_S \rangle = c_0 - P_S^2(t_i)$ achieves the maximum $c_0$ precisely on $S$. The construction requires one dimension per coefficient from $t^1$ through $t^{2k}$, yielding the $2k$ upper bound.

\begin{theorem}\label{thm:linear-lower-upper}
    $$k-1\leq \med(m, k; \fln) \leq 2k.$$
\end{theorem}
\begin{proof}
The lower bound follows from the VC dimension of $\fln$ in $\real^n$ being $n+1$~\citep{mohri2018foundations} together with Proposition~\ref{prop:vc-bounds}, giving $\med(m,k;\fln) \ge k-1$. For the upper bound, take $n = 2k$ and embed $m$ points on the moment curve with distinct parameters. Lemma~\ref{lem:poly-construction} provides, for every $S\subseteq X$ with $|S|\le k$, a query vector $\vq_S$ that strictly separates $S$ from $X\setminus S$ by inner product. Hence $m$ points in $\real^{2k}$ are $k$-shattered by $\fln$, and $\med(m,k;\fln) \le 2k$.
\end{proof}

For completeness, Appendix~\ref{app:radon-linear-tightening} records a Radon-theorem refinement showing that the inner-product bound above can be sharpened to
\(\mathrm{MED}(m,k;F_{\rm linear})=\min\{2k,m-1\}\), while the main text keeps the \(\Theta(k)\) presentation because the Euclidean and cosine transfers below are used only at the displayed-bound level.
We see that $\med(m, k; \fln)$ only depends on $k$. Ignoring the coefficient, $\med(m, k; \fln) = \Theta(k)$ also holds. Then, we show that $\fcos$ and $\fl{2}$ share similar bounds as $\fln$.

\subsection{Euclidean-score bounds}~\label{sec:generalize_linear}

By geometric constructions in $\real^n$ regarding the $k$-shattering, the following relation is revealed.

\begin{proposition}\label{prop:l2<linear}

$$\med(m, k; \fl{2}) \leq \med(m, k; \fln). $$

\end{proposition}
\begin{proof}
Suppose $X\subseteq\real^n$ is $k$-shattered by $\fln$. For $S=X$, the required separation is vacuous, so assume $X\setminus S$ is nonempty. For each such $S\in\collection_k$, choose a separator $\vq_S$ and threshold $b_S$ such that $\langle \vq_S,\vx\rangle>b_S>\langle \vq_S,\vy\rangle$ for all $\vx\in S$ and $\vy\notin S$. Since $X$ is finite, the margin
\[
    \Delta_S=\min_{\vx\in S,\ \vy\notin S}
    \langle \vq_S,\vx-\vy\rangle
\]
is positive. Let $\vu_S=\vq_S/\|\vq_S\|_2$. For $R>0$, set the Euclidean query center to $\vw_S=R\vu_S$. Then
\[
    \|\vy-\vw_S\|_2^2-\|\vx-\vw_S\|_2^2
    =
    2R\langle \vu_S,\vx-\vy\rangle+\|\vy\|_2^2-\|\vx\|_2^2 .
\]
Choosing $R$ large enough makes this quantity positive for every selected--unselected pair. Thus every selected point is strictly closer to $\vw_S$ than every unselected point, so the same configuration is $k$-shattered by $\fl{2}$.
\end{proof}
When considering the Proposition~\ref{prop:l2<linear}, we conclude $\med(m, k;\fl{2})$ in Theorem~\ref{thm:l2-lower-upper} with additional information that $\vcd(n;\fl{2}) = n+1$ and Proposition~\ref{prop:vc-bounds}.

\begin{theorem}\label{thm:l2-lower-upper}
    $$k-1\leq \med(m, k; \fl{2}) \leq 2k.$$
\end{theorem}

\subsection{Cosine-score bounds}

\begin{proposition}\label{prop:linear=cos}
\begin{align*}
    \med(m, k; \fln) &\leq \med(m, k; \fcos), \\
    \med(m, k; \fcos) &\leq \med(m, k; \fln) + 1.
\end{align*}
\end{proposition}
\noindent\textit{Proof sketch:} A cosine query on nonzero objects depends only on the normalized directions $x/\|x\|_2$ and a threshold, so any cosine-shattered configuration gives a linearly shattered configuration on the unit sphere in the same dimension. Conversely, any affine linear separator $\langle \vw_S,x\rangle>b_S$ can be homogenized by mapping $x\mapsto (x,1)/\|(x,1)\|_2$ and using the cosine query direction $(\vw_S,-b_S)$. This costs one extra dimension. The full proof is given in Appendix~\ref{app:proof:prop:linear=cos}.

Combination of the Theorem~\ref{thm:linear-lower-upper} and the Proposition~\ref{prop:linear=cos} also shows that $\med(m, k; \fcos) = \Theta(k)$.

\begin{theorem}\label{thm:cos-lower-upper}
    $$k-1\leq \med(m, k; \fcos) \leq 2k+1.$$
\end{theorem}

\subsection{Discussion on exact MED}
To summarize, the minimum embeddable dimensions for $\fln$, $\fcos$, and $\fl{2}$ are all $\Theta(k)$, independent of the number of total elements in the set system up to constants. The inner-product and Euclidean constructions use at most $2k$ dimensions, while the cosine construction uses at most $2k+1$ dimensions. The exact centroid-query upper bounds for the three scoring rules follow by dropping the positive margin from the robust centroid-query construction in Proposition~\ref{prop:rmedc-consequences-main}, using Proposition~\ref{prop:centroid-witnesses}.

The bounds also clarify what exact separability can and cannot explain. The geometric score family does not limit exact threshold retrieval for answer sets of size at most $k$: one can put the elements on the moment curve in $\real^{2k}$ and use the squared-polynomial query vector above for each target set. This proves the existence of exact low-dimensional witnesses, not a deployable query encoder that can generalize. Practical failures may still arise from positive margins, conditioning, finite precision, or optimization protocol. The robust setting in Section~\ref{sec:rmed-theory} shows that the dimension regime changes from $\Theta(1)$ to $O(k^2\log m)$ (when the margin $\epsilon$ is proper) once a normalized score gap is required.
\section{Robust MED: Margin Regimes}\label{sec:rmed-theory}

The previous section studies exact separability: the selected objects only need to score strictly above the unselected objects. This section first studies the normalized inner-product version $\rmed(m,k,\epsilon;\fln)$ from Definition~\ref{def:rmed} in the positive-margin regime $\epsilon>0$, where all object and query vectors are normalized and every top-$k$ query must have score gap at least $\epsilon$. Unless stated otherwise, finite-$m$ RMED statements keep $\epsilon_\star(m,k)$ and use $1\leq k\leq m/2$; asymptotic phrases such as the $1/\sqrt{k}$ margin scale refer to the large-universe regime $m/k\to\infty$. Exact MED is the no-margin problem; it should not be identified with the literal limit $\epsilon=0$ under the non-strict robust inequality.

\subsection{Lower bounds}

The first lower bound follows directly from MED. If $\epsilon$-robust $k$-shattering is possible in dimension $d$ for $\epsilon>0$, then exact $k$-shattering is also possible in the same dimension. Therefore,
\begin{align}
    \med(m,k;\fln) \leq \rmed(m,k,\epsilon;\fln)
    \label{eq:med-rmed-relation}
\end{align}
whenever $\epsilon>0$ and $\rmed(m,k,\epsilon;\fln)$ is finite. In particular, $\rmed(m,k,\epsilon;\fln)\geq \Omega(k)$ by Theorem~\ref{thm:linear-lower-upper}.

A positive normalized margin also forces a packing lower bound. \citet{weller2025theoretical} prove that if all $k$-subsets can be realized with margin $\gamma$ in their notation, then
\[
    \binom{m}{k} \leq \left(1+\frac{1}{\gamma}\right)^d .
\]
Their margin $\gamma$ corresponds to a score gap $\epsilon=2\gamma$ in Definition~\ref{def:robust-shattering}. Thus,
\begin{theorem}[Packing lower bound]\label{thm:rmed-packing}
For $1\leq k\leq m/2$ and $\epsilon>0$,
\begin{align}
    \rmed(m,k,\epsilon;\fln)
    \geq
    \frac{\log \binom{m}{k}}{\log(1+2/\epsilon)}.
    \label{eq:rmed-packing}
\end{align}
\end{theorem}
When $k\leq m/2$, this gives $\Omega(k\log(em/k)/\log(1+2/\epsilon))$. The $\Omega(k)$ lower bound in \eqref{eq:med-rmed-relation} should be kept separately, especially when $\epsilon$ is very small.

\subsection{Feasibility ceiling}

Before asking how large $d$ must be, one must ask whether the requested margin is feasible at all.

\begin{theorem}[Margin feasibility ceiling]\label{thm:rmed-feasibility}
If $\rmed(m,k,\epsilon;\fln)<\infty$ for $1\leq k\leq m/2$, then
\begin{align}
    \epsilon
    \leq
    \epsilon_\star(m,k)
    :=
    \frac{m}{\sqrt{k(m-1)(m-k)}}.
    \label{eq:eps-star}
\end{align}
In particular, in the large-universe regime $m/k\to\infty$, $\epsilon_\star(m,k)\sim 1/\sqrt{k}$.
\end{theorem}

\begin{proof}
Assume unit vectors $\vv_1,\dots,\vv_m$ and unit queries $\vu_S$ realize every $k$-subset with gap $\epsilon$. For any $S$ with $|S|=k$, every $i\in S$ and $j\notin S$ satisfy
\[
    \langle \vu_S,\vv_i-\vv_j\rangle\geq \epsilon.
\]
Averaging over $i\in S$ and $j\notin S$ gives
\[
    \left\langle \vu_S,
    \bar{\vv}_S-\bar{\vv}_{S^c}
    \right\rangle\geq \epsilon,
\]
where $\bar{\vv}_S=\frac{1}{k}\sum_{i\in S}\vv_i$ and $\bar{\vv}_{S^c}=\frac{1}{m-k}\sum_{j\notin S}\vv_j$. Hence $\|\bar{\vv}_S-\bar{\vv}_{S^c}\|_2\geq \epsilon$.
Let $\bar{\vv}=\frac{1}{m}\sum_i \vv_i$. Since
\[
    \bar{\vv}_S-\bar{\vv}_{S^c}
    =
    \frac{m}{m-k}(\bar{\vv}_S-\bar{\vv}),
\]
we have $\|\bar{\vv}_S-\bar{\vv}\|_2\geq \frac{m-k}{m}\epsilon$ for every $k$-subset $S$.

Now choose $S$ uniformly among all $k$-subsets. The finite-population variance identity gives
\[
    \mathbb{E}_S\|\bar{\vv}_S-\bar{\vv}\|_2^2
    =
    \frac{m-k}{k(m-1)}
    \cdot
    \frac{1}{m}\sum_{i=1}^m \|\vv_i-\bar{\vv}\|_2^2
    \leq
    \frac{m-k}{k(m-1)},
\]
because the vectors are unit norm. Therefore some $S$ satisfies
\[
    \frac{m-k}{m}\epsilon
    \leq
    \sqrt{\frac{m-k}{k(m-1)}} ,
\]
which is equivalent to \eqref{eq:eps-star}.
\end{proof}

Thus a fixed positive score gap cannot survive as $k$ grows. For $k\leq m/2$, $\epsilon_\star^2(m,k)=k^{-1}\cdot m/(m-1)\cdot m/(m-k)$, so $\epsilon_\star(m,k)\asymp1/\sqrt{k}$ and converges to $1/\sqrt{k}$ when $m/k\to\infty$. This feasibility limit is compatible with exact MED, which imposes no prescribed normalized margin, while feasible robust retrieval still depends on $m$ through the packing lower bound and the Gaussian centroid witness. The ceiling is sharp in dimension $m-1$ by a regular simplex construction; the proof and a Johnson--Lindenstrauss compression are given in Appendix~\ref{app:rmed-simplex-jl}.

\subsection{Upper bounds}

The Gaussian centroid construction gives a simple upper bound at the natural margin scale.

\begin{theorem}[Gaussian centroid robust witness]\label{thm:rmed-gaussian}
There exists a universal constant $c>0$ such that, for $2\leq k\leq m/2$ and $\epsilon_k=c/\sqrt{k}$,
\begin{align}
    \rmedc(m,k,\epsilon_k;\fln)
    &\leq
    O(k^2\log m).
    \label{eq:rmed-gaussian}
\end{align}
Consequently, $\rmed(m,k,\epsilon_k;\fln)\leq O(k^2\log m)$.
\end{theorem}

\noindent\textit{Proof sketch.}
Sample $m$ vectors from an isotropic Gaussian distribution and normalize them. With positive probability in dimension $n=Ck^2\log m$, all pairwise inner products are $O(1/k)$, and all norms concentrate near one. Positive probability is sufficient for the existential upper bound; by increasing the universal constant $C$, the finite union-bound failure probability can be made arbitrarily small. For each $S$, take the query direction to be the normalized centroid $\vu_S\propto \sum_{i\in S}\vv_i$. A selected object has a self-correlation term of order one, while every outsider contributes only coherence noise of order $|S|/k$. After normalizing the centroid, whose norm is $\Theta(\sqrt{|S|})$, the inner-product score gap is $\Omega(1/\sqrt{k})$ uniformly over all $S\in\collection_k$. Thus the construction witnesses the restricted quantity $\rmedc(\cdot;\fln)$, and therefore also witnesses $\rmed(\cdot;\fln)$. The full normalized-margin calculation is given in Appendix~\ref{app:proof:thm:rmed-gaussian}.

The same unit-vector construction transfers across the scoring rules used in this paper. Cosine scores agree with inner products after normalization, so the same bound holds for $\rmed(\cdot;\fcos)$ and $\rmedc(\cdot;\fcos)$. For Euclidean scoring, comparing squared distances to a shared unit query gives
\[
    \|\vv_j-\vu_S\|_2^2-\|\vv_i-\vu_S\|_2^2
    =
    2\langle \vu_S,\vv_i-\vv_j\rangle ,
\]
so an inner-product margin $\epsilon$ gives a squared-distance gap $2\epsilon$ and an ordinary distance gap at least $\epsilon/2$. Hence the same construction also gives $\rmed(m,k,c'/\sqrt{k};\fl{2})\leq O(k^2\log m)$ for another universal constant $c'>0$.

We record the robust counterpart of the centroid-query consequence before dropping the margin.
\begin{proposition}[Robust centroid-query consequences]\label{prop:rmedc-consequences-main}
There exist universal constants $c_1,c_2>0$ such that, for $2\leq k\leq m/2$,
\begin{align*}
    \rmedc(m,k,c_1/\sqrt{k};\fln)&\leq O(k^2\log m),\\
    \rmedc(m,k,c_1/\sqrt{k};\fcos)&\leq O(k^2\log m),\\
    \rmedc(m,k,c_2/\sqrt{k};\fl{2})&\leq O(k^2\log m).
\end{align*}
\end{proposition}
\noindent\textit{Proof.} See Appendix~\ref{app:proof:prop:rmedc-consequences-main}.

The centroid upper bound should be read with its remaining gap in mind.
The exact MED lower bound gives only an inherited $\Omega(k)$ lower
bound for centroid-query witnesses, while the Gaussian construction
uses $O(k^2\log m)$ dimensions. The logarithmic dependence on $m$
already separates this witness family from the cubic fitted baseline in
Section~\ref{sec:maed_and_exp}, but the extra factor in $k$ may not be
tight. Related sparse one-bit recovery problems can achieve
$k\log(m/k)$-type sample complexity only with additional recovery
structure~\citep{aksoylar2014information}, suggesting that improving
centroid-query bounds may require more than independent score
comparison.

The zero-margin centroid-query MED upper bounds are the exact consequences of this robust centroid construction after dropping the margin, as noted in Section~\ref{sec:optimal-standard-theory}.

\subsection{Regime summary}

For finite $m,k$ with $k\leq m/2$, the theory separates the exact and robust regimes:
\begin{compactitem}
    \item Exact MED is the separate no-margin problem: it has lower bound $\Omega(k)$ and upper bound $2k+O(1)$.
    \item $0<\epsilon<\epsilon_\star(m,k)$: positive-margin inner-product RMED is subject to the lower bounds \eqref{eq:med-rmed-relation} and \eqref{eq:rmed-packing}; Appendix~\ref{app:rmed-simplex-jl} records a high-dimensional sharpness construction for the feasibility ceiling.
    \item $\epsilon=c/\sqrt{k}$: for a sufficiently small universal $c$, the Gaussian centroid construction gives $O(k^2\log m)$ dimensions.
    \item $\epsilon>\epsilon_\star(m,k)$: the robust problem is infeasible.
\end{compactitem}
The main message is that exact approximability is low-dimensional, while robust normalized approximability has a different regime. Exact top-$k$ embeddability is governed by neighborliness and costs $\Theta(k)$ dimensions, whereas robust normalized retrieval is governed by packing and averaging: positive margin makes $m$ reappear in the dimension bounds, and score gaps above $\epsilon_\star(m,k)$ are infeasible. Under the large-universe RMED convention $m/k\to\infty$, this ceiling is asymptotic to $1/\sqrt{k}$.
\section{Numerical simulation and implications}~\label{sec:maed_and_exp}

This section asks whether the empirical evidence points to exact-approximability failures. It does not, in fact. \footnote{Source code for experiments are published in \url{https://github.com/zihao-wang/med}}

\subsection{Synthetic top-2 test set optimization}
The cyclic-polytope construction certifies exact MED upper-bound witnesses, while mean-embedding optimization probes the centroid/Gaussian regime from Theorem~\ref{thm:rmed-gaussian}. We optimize $m$ embeddings with a hinge loss over all positive-negative pairs using Adam~\citep{kingma2014adam}; cyclic-polytope and LIMIT artifacts are generated deterministically from the packaged data. A zero-violation run certifies that the checked dimension supports the tested constraints, but a failed lower-dimensional run is not a proof of infeasibility.

\begin{figure}[t]
  \centering
  \includegraphics[width=.8\linewidth]{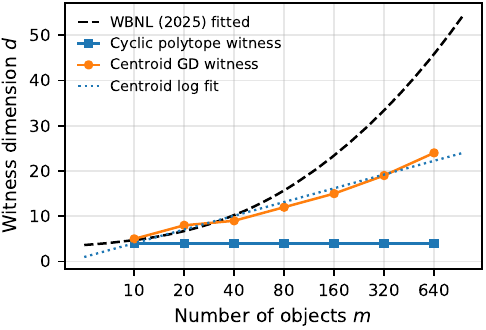}
  \caption{Synthetic top-$2$ witness dimension as a function of universe size. The cyclic-polytope construction stays at dimension $4$, while centroid GD witnesses grow slowly in $\log_2 m$; the WBNL fitted curve is included as a reference, translated from their polynomial fitting of maximum $m$ given $d$.}
  \label{fig:top2-synthetic}
\end{figure}

For comparison, \citet{weller2025theoretical} fit a cubic curve for the largest successfully optimized universe size in dimension $d$ when $k=2$. We translated this curve into our MED setting to find the witness of dimensions that successfully embedded $m$ objects, which is called WBNL.
Figure~\ref{fig:top2-synthetic} shows the synthetic top-$2$ witness dimension as a function of universe size. The cyclic-polytope construction gives an exact dimension-$4$ witness for every tested $m$, while the centroid GD witness and its log-linear fit remain far below the dimension predicted by the fitted WBNL curve on this grid. These plotted points should be read as checked upper-bound witnesses rather than certified minima. It suggests that the claim in~\citep{weller2025theoretical}, as quoted, ``for web-scale search, even the largest embedding dimensions with ideal test-set optimization are not enough to model all combinations'' is questionable.

\citet{weller2025theoretical} failed without contradicting the MED result. MED is an exact-approximability statement: it asks for the minimal dimension where suitable object embeddings and query vectors exist. The WBNL curve tells when their free-embedding optimization protocol learns such vectors, which is also a witness upper bound. This also clarifies why the optimization protocol is essential in embedding-based retrieval experiments compared with the centroid query experiments.
If both object vectors and all subset query vectors were optimized to global optimality, the free-embedding hinge objective could only improve on a centroid-query objective, because the centroid-query setting has the centroid solution as a special constrained case. Observing weaker points from the free protocol therefore says more about the optimization landscape and query-learning protocol than about the exact approximability of top-$2$ set systems.

\subsection{LIMIT retrieval: suprising power of single-vector embeddings}

We further evaluate the LIMIT and LIMIT-small datasets from~\citet{weller2025theoretical} using label-unaware single-vector \textbf{random additive construction}. For a tokenizer $\tau$, each observed token $t$ receives a fixed unit Gaussian vector in $G_t\in \mathbb{R}^d$; a document or query $x$ is embedded as $\phi(x)=\sum_{t\in\tau(x)}G_t$; and documents are ranked by the inner product $\langle \phi(q),\phi(c)\rangle$. Thus, the comparison remains a single-vector retrieval comparison on the same Recall@2 metric. We study three kinds of tokenizers: qwen for \texttt{Qwen3-0.6B} tokenizer downloaded from huggingface, vanilla for tokenization by space or punctuation with stop words removed, and handmade, which leverages the vocabulary to build the LIMIT and LIMIT-small datasets. We use the vanilla tokenizer to justify our argument. Qwen and handmade tokenizers are used for comparison. Appendix~\ref{app:random-token-limit} gives the algorithm and the full Recall@2 grid.

\begin{figure}[t]
  \centering
  \begin{subfigure}[t]{0.49\linewidth}
    \centering
    \includegraphics[width=\linewidth]{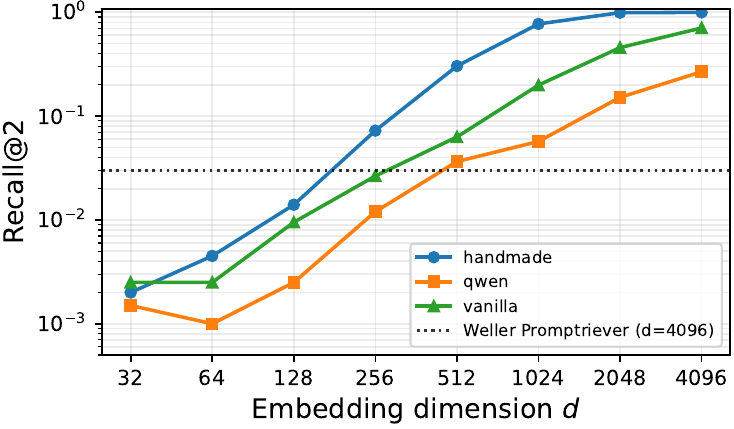}
    \caption{LIMIT}
  \end{subfigure}\hfill
  \begin{subfigure}[t]{0.49\linewidth}
    \centering
    \includegraphics[width=\linewidth]{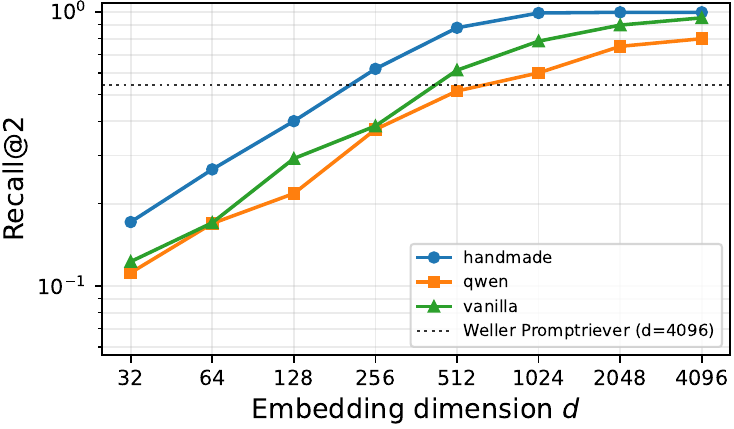}
    \caption{LIMIT-small}
  \end{subfigure}
  \caption{Recall@2 across embedding dimensions and constructions.
  Dotted horizontal lines mark Weller et al.'s 4096-dimensional Promptriever Llama3 8B single-vector embedding scores.}
  \label{fig:limit-retrieval}
\end{figure}

The dotted lines in Figure~\ref{fig:limit-retrieval} use the strongest comparable single-vector embedding result reported by \citet{weller2025theoretical}: Promptriever Llama3 8B at dimension $4096$, with Recall@2 equal to $0.030$ on LIMIT and $0.543$ on LIMIT-small. We do not use BM25, tokenwise TF-IDF, GTE-ModernColBERT, or Gemini long-context reranking as baselines for this capacity comparison because they are sparse, tokenwise, multi-vector, cross-encoder, or long-context reranking methods rather than single-vector embedding baselines.

All three tokenizations cross the Promptriever line. On full LIMIT, the handmade tokenizer crosses by $d=256$, while Qwen and vanilla tokenizers cross by $d=512$; at $d=4096$, their Recall@2 values are $0.9980$, $0.2675$, and $0.7060$, respectively, versus the Promptriever value $0.030$. On LIMIT-small, the corresponding crossing dimensions are $d=256$, $1024$, and $512$, and the $d=4096$ Recall@2 values are $1.0000$, $0.8010$, and $0.9545$, respectively, versus the Promptriever value $0.543$. The vanilla tokenizer is the most important control for interpretation: it uses neither handcrafted phrases nor LLM subwords, and it is not supervised on LIMIT labels, yet it still exceeds the Promptriever baseline by a wide margin. For performance, the handmade tokenizer is especially favorable when prior knowledge is given so that we know the phrases to be retrieved in advance. Qwen tokenizer performed weakly in this setting, suggesting it has collaborated with token embeddings and pretrained LMs for more sophisticated information aggregation.

These results expose an identification problem in interpreting the LIMIT experiments of \citet{weller2025theoretical}. Random token sums with a vanilla tokenizer might be the simplest model we can imagine. The failure of a learned single-vector encoder on LIMIT, reported by~\citep{weller2025theoretical} is not evidence that dimension $4096$ lacks exact single-vector geometric capacity.  Poor learned-embedding performance can also reflect learnability, feature allocation, tokenizer geometry, objective choice, margin conditioning, or finite-precision effects.

Separately, the cyclic-polytope construction can exactly overfit the packaged LIMIT and LIMIT-small top-$2$ datasets in dimension $4$. Appendix~\ref{app:cyclic-limit} reports the summary and exported LIMIT-small document and query vectors. Together with this exact cyclic-polytope overfit, the random token-sum controls show that the LIMIT failures are not exact-approximability failures. They are better understood as failures or stresses of learned representations, tokenization, objectives, margin conditioning, or optimization protocol.
\section{Related Theoretical Works}\label{sec:related}

We focus on theoretical comparisons most likely to be confused with MED, rather than surveying empirical retrieval systems.

\citet{weller2025theoretical} study a positive-margin version of embedding retrieval: for every $k$-subset $S$, selected objects must score at least $t+g$ and unselected objects at most $t-g$. In their corpus-size notation, this gives a lower bound of order
\[
    \frac{k\log(m/k)}{\log(1+1/g)}
\]
For constant-order margins. This is compatible with our robust MED result. In their original version, a sign-rank upper bound is given, which is clearly over-complex compared to the construction raised in this paper. Their free-embedding optimization experiments also test whether a particular learning protocol finds witnesses; failure of that protocol is evidence about optimization, not a geometric lower bound for exact embeddability.

\citet{guo2019breaking} study multi-class embedding classification and prove dimension bounds such as $O(\min\{s\log(K|S|),s^2\log K\})$ in their notation. Translating arbitrary at-most-$k$ answer sets gives $\sum_{i=1}^k\binom{m}{i}$ possible classes, much larger than the structured classes used in their application. \citet{you2025hierarchical} study hierarchical retrieval where answer sets are reachable sets in a directed acyclic graph; that family is tied to the source objects and graph structure, whereas MED contains all arbitrary answer sets up to size $k$. We therefore treat both as related structured settings, not as direct upper or lower bounds for arbitrary top-$k$ MED. Appendix~\ref{app:additional-comparisons} gives the longer sign-rank comparison and corrected parameter translations for these structured settings.
\section{Conclusion}\label{sec:conclusion}

This paper answers the question of the minimal embeddable dimension for top-$k$ retrieval. Exact MED has a favorable answer: cyclic-polytopal neighborliness gives $\Theta(k)$ dimensions for inner product, cosine, and Euclidean scoring, independent of $m$ up to constants. Robust MED is different: finite-$m$ score gaps are capped by $\epsilon_\star(m,k)$, and in the large-universe feasible $c/\sqrt{k}$ regime the Gaussian centroid witness gives $O(k^2\log m)$ dimensions. Thus, the empirical failures considered here are not failures of geometric approximability; the remaining difficulties lie in learning, tokenization, objectives, conditioning, finite precision, and optimization.
\section*{Impact Statement}

This paper studies minimal dimensions for embedding-based retrieval,
highlighting a theoretical optimum that may help guide
retrieval-system development. This is a highly theory-focused task
that mainly contains mathematical propositions and proofs. No human
subjects were involved in this work. However, this work may still
lead to unexpected negative societal impacts, depending on practical
developments that we cannot foresee at this stage.

\section*{Acknowledgement}

The authors want to thank anonymous reviewers and area chairs in the ICLR'26 and the ICML'26 submission process and Professor Yury Polyanskiy for their feedback.

\bibliography{preprint}
\bibliographystyle{icml2026}

\newpage
\appendix
\onecolumn
\clearpage
\begin{figure}[t!]
  \huge
  \begin{center}
    \textbf{Supplementary Material}
  \end{center}
\end{figure}

\section{Additional Proofs}
\subsection{Radon sharpening of the linear exact MED bound}
\label{app:radon-linear-tightening}

This appendix records an optional sharpening of the inner-product
lower bound in Theorem~3.3. The main text only needs the order-level
\(\Theta(k)\) statement, but the linear case itself admits an exact
constant by Radon's theorem.

\begin{theorem}[Radon's theorem]
\label{thm:radon}
Let \(Y\subset \mathbb R^d\) be a set of \(d+2\) points. Then there
exists a partition
\[
    Y=A\dot\cup B,
    \qquad A,B\neq \emptyset,
\]
such that
\[
    \operatorname{conv}(A)\cap \operatorname{conv}(B)\neq \emptyset .
\]
\end{theorem}

\begin{proposition}[Exact linear MED]
\label{prop:radon-linear-med}
For \(m\ge 2\) and \(1\le k\le m\),
\[
    \mathrm{MED}(m,k;F_{\rm linear})
    =
    \min\{2k,m-1\}.
\]
In particular, when \(m\ge 2k+1\),
\[
    \mathrm{MED}(m,k;F_{\rm linear})=2k.
\]
\end{proposition}

\begin{proof}
We first prove the upper bound. Lemma~3.2 gives a moment-curve
construction in dimension \(2k\), hence
\[
    \mathrm{MED}(m,k;F_{\rm linear})\le 2k .
\]
Also, \(m\) affinely independent points in \(\mathbb R^{m-1}\) can
realize every subset by affine thresholding. Indeed, for any subset
\(S\subseteq [m]\), there is an affine functional \(h_S\) that takes
value \(1\) on the vertices indexed by \(S\) and value \(0\) on the
remaining vertices; thresholding at \(1/2\) strictly separates \(S\)
from its complement. Since affine thresholding is equivalent to a
linear score with a threshold, this gives
\[
    \mathrm{MED}(m,k;F_{\rm linear})\le m-1 .
\]
Therefore
\[
    \mathrm{MED}(m,k;F_{\rm linear})
    \le
    \min\{2k,m-1\}.
\]

It remains to prove the lower bound. Suppose, for contradiction, that
some configuration
\[
    X=\{x_1,\ldots,x_m\}\subset \mathbb R^d
\]
is \(k\)-shattered by \(F_{\rm linear}\) in a dimension satisfying
\[
    d < \min\{2k,m-1\}.
\]
Since \(d<m-1\), we have \(d+2\le m\), so we may choose a subset
\(Y\subset X\) with \(|Y|=d+2\). By Radon's theorem, there is a
partition
\[
    Y=A\dot\cup B,
    \qquad A,B\neq \emptyset,
\]
such that
\[
    \operatorname{conv}(A)\cap \operatorname{conv}(B)\neq \emptyset .
\]

Because \(d<2k\), we have \(d+2\le 2k+1\). Hence one side of the
Radon partition has size at most \(k\). Without loss of generality,
rename the two sides so that
\[
    1\le |A|\le k .
\]
Since \(X\) is \(k\)-shattered, there exist \(w\in\mathbb R^d\) and
\(b\in\mathbb R\) such that
\[
    \langle w,a\rangle > b > \langle w,x\rangle
    \qquad
    \forall a\in A,\ \forall x\in X\setminus A .
\]
In particular, this strict separation holds between \(A\) and \(B\).
By convexity, every point in \(\operatorname{conv}(A)\) has inner
product strictly larger than \(b\), while every point in
\(\operatorname{conv}(B)\) has inner product strictly smaller than
\(b\). Therefore
\[
    \operatorname{conv}(A)\cap \operatorname{conv}(B)=\emptyset,
\]
contradicting Radon's theorem.

Thus no dimension
\[
    d<\min\{2k,m-1\}
\]
can \(k\)-shatter \(m\) points under \(F_{\rm linear}\). Hence
\[
    \mathrm{MED}(m,k;F_{\rm linear})
    \ge
    \min\{2k,m-1\}.
\]
Combining the upper and lower bounds proves the claim.
\end{proof}

\subsection{Proof of Proposition~\ref{prop:linear=cos}}\label{app:proof:prop:linear=cos}

\begin{proof}
First suppose $X=\{x_1,\dots,x_m\}\subset\real^n$ is $k$-shattered by cosine similarity. Cosine scores require nonzero object vectors; moreover, if two objects had the same normalized direction, singleton queries could not distinguish them. Define
\[
    \rho(x)=\frac{x}{\|x\|_2},
    \qquad
    Y=\{\rho(x_i)\}_{i=1}^m\subset S^{n-1}.
\]
For $S=X$, both exact shattering conditions are vacuous on the unselected side, so assume $X\setminus S$ is nonempty. For each such $S\in\collection_k$, cosine shattering gives a nonzero query $\vw_S$ and threshold $b_S$ such that
\[
    \frac{\langle \vw_S,x\rangle}{\|\vw_S\|_2\|x\|_2}
    > b_S
    \quad (x\in S),
    \qquad
    \frac{\langle \vw_S,y\rangle}{\|\vw_S\|_2\|y\|_2}
    < b_S
    \quad (y\notin S).
\]
Equivalently, the normalized points $Y$ are separated by the linear functional
$z\mapsto\langle \vw_S/\|\vw_S\|_2,z\rangle$ with threshold $b_S$. Hence
$\med(m,k;\fln)\leq \med(m,k;\fcos)$.

Conversely, suppose $X=\{x_1,\dots,x_m\}\subset\real^n$ is $k$-shattered by $\fln$. For each $S\in\collection_k$ with $X\setminus S$ nonempty, choose $\vw_S$ and $b_S$ such that
\[
    \langle \vw_S,x\rangle>b_S
    \quad (x\in S),
    \qquad
    \langle \vw_S,y\rangle<b_S
    \quad (y\notin S).
\]
Embed the objects on the unit sphere in one higher dimension by
\[
    \phi(x)=\frac{(x,1)}{\|(x,1)\|_2}\in S^n\subset\real^{n+1}.
\]
For the query direction, take
\[
    \vu_S=\frac{(\vw_S,-b_S)}{\|(\vw_S,-b_S)\|_2},
\]
with any nonzero direction chosen in the trivial case $(\vw_S,-b_S)=0$.
Then
\[
    \langle \vu_S,\phi(x)\rangle
    =
    \frac{\langle \vw_S,x\rangle-b_S}
    {\|(\vw_S,-b_S)\|_2\|(x,1)\|_2}.
\]
The denominator is positive, so threshold zero separates $\phi(S)$ from
$\phi(X\setminus S)$ by cosine similarity. Thus
$\med(m,k;\fcos)\leq \med(m,k;\fln)+1$.
\end{proof}

\subsection{Simplex sharpness and JL compression}\label{app:rmed-simplex-jl}

\begin{proposition}[Simplex tightness]\label{prop:rmed-simplex}
For $1\leq k\leq m/2$, $m$ vertices of a regular simplex in $\real^{m-1}$ realize every nonempty subset of size at most $k$ with score gap at least $\epsilon_\star(m,k)$. Consequently, every $\epsilon\leq \epsilon_\star(m,k)$ is feasible in dimension $m-1$.
\end{proposition}

\begin{proof}
Let $\vv_1,\dots,\vv_m$ be unit vectors with $\langle \vv_i,\vv_j\rangle=-1/(m-1)$ for $i\ne j$. For $|S|=k$, choose
\[
    \vu_S=\frac{\sum_{i\in S}\vv_i}{\left\|\sum_{i\in S}\vv_i\right\|_2}.
\]
For $i\in S$,
\[
    \left\langle \sum_{\ell\in S}\vv_\ell,\vv_i\right\rangle
    =
    \frac{m-k}{m-1},
\]
while for $j\notin S$,
\[
    \left\langle \sum_{\ell\in S}\vv_\ell,\vv_j\right\rangle
    =
    -\frac{k}{m-1}.
\]
The unnormalized score gap is $m/(m-1)$, and
\[
    \left\|\sum_{i\in S}\vv_i\right\|_2^2
    =
    \frac{k(m-k)}{m-1}.
\]
After query normalization, the selected--unselected score gap is
\[
    \frac{m/(m-1)}{\sqrt{k(m-k)/(m-1)}}
    =
    \frac{m}{\sqrt{k(m-k)(m-1)}}
    =
    \epsilon_\star(m,k),
\]
which is exactly the feasibility ceiling from Theorem~\ref{thm:rmed-feasibility}. Thus the upper bound on $\epsilon$ is attained by the regular simplex in dimension $m-1$. For a subset of size $s\leq k$, the same calculation gives gap $\epsilon_\star(m,s)$. When $k\leq m/2$, $\epsilon_\star(m,s)\geq \epsilon_\star(m,k)$ for every $1\leq s\leq k$.
\end{proof}

The simplex construction is margin-optimal but has dimension $m-1$. A standard Johnson--Lindenstrauss projection compresses the construction while preserving the relevant query-object inner products.

\begin{theorem}[Simplex plus JL upper bound]\label{thm:rmed-jl}
Assume $1\leq k\leq m/2$ and $0<\epsilon<\epsilon_\star(m,k)$. Then
\begin{align}
    \rmed(m,k,\epsilon;\fln)
    \leq
    O\!\left(
    \frac{k\log(em/k)}
    {\max(1, (\epsilon_\star(m,k)-\epsilon)^2)}
    \right).
    \label{eq:rmed-jl}
\end{align}
\end{theorem}

\begin{proof}
Start from the simplex construction and collect the $m$ object vectors and all normalized query vectors $\vu_S$ for $1\leq |S|\leq k$. The number of vectors is
\[
    N=m+\sum_{s=1}^k \binom{m}{s},
\]
so $\log N=O(k\log(em/k))$ for $k\leq m/2$. By Johnson--Lindenstrauss inner-product preservation~\citep{vershynin2018high}, a random projection to dimension $O(\log N/\delta^2)$ preserves all query-object inner products up to additive error $\delta$. Choosing $\delta$ as a sufficiently small constant multiple of $\epsilon_\star(m,k)-\epsilon$ leaves score gap at least $\epsilon$, giving \eqref{eq:rmed-jl}.
\end{proof}

\subsection{Proof of Theorem~\ref{thm:rmed-gaussian}}\label{app:proof:thm:rmed-gaussian}

We prove the Gaussian centroid robust witness. The zero-margin MED-C consequences are recorded in Appendix~\ref{app:medc-consequences}.

\begin{proof}
Let
\[
    n=\lceil Ck^2\log m\rceil
\]
for a universal constant $C$ chosen below. Sample independent Gaussian
vectors $g_1,\dots,g_m\sim\mathcal{N}(0,I_n)$ and normalize them by
\[
    \vv_i=\frac{g_i}{\|g_i\|_2}.
\]
The normalization puts every object on the unit sphere; the only
probabilistic property needed below is pairwise near-orthogonality. We
first spell out the concentration inequality that gives this property.

Fix a pair $i\ne j$ and condition on $\vv_i$. By rotational invariance
of the Gaussian distribution, we may rotate coordinates so that
$\vv_i=e_1$. If $g\sim\mathcal{N}(0,I_n)$ is independent, then
$\vv_j$ has the same distribution as $g/\|g\|_2$, and therefore
\[
    \langle \vv_i,\vv_j\rangle
    \stackrel{d}{=}
    \frac{g_1}{\|g\|_2}.
\]
For $0<t\le 1/2$, the event $|g_1|/\|g\|_2>t$ can occur only if
$|g_1|>t\sqrt{n/2}$ or if $\|g\|_2^2<n/2$. The Gaussian tail bound and
the lower tail bound for a chi-square random variable therefore imply
that, for universal constants $a_0,a_1>0$,
\[
\begin{aligned}
    \mathbb{P}\left(|\langle \vv_i,\vv_j\rangle|>t\right)
    &\le
    \mathbb{P}\left(|g_1|>t\sqrt{n/2}\right)
    +\mathbb{P}\left(\|g\|_2^2<n/2\right) \\
    &\le
    2\exp(-a_0nt^2)+\exp(-a_1n)
    \le
    3\exp(-a_0nt^2),
\end{aligned}
\]
after decreasing $a_0$ if necessary. This is the usual spherical-cap
concentration inequality: every fixed one-dimensional projection of a
random unit vector is sub-Gaussian at scale $1/\sqrt{n}$.

Set $t=1/(4k)$. Since $k\ge2$, this lies in the range above. A union
bound over fewer than $m^2/2$ unordered pairs gives
\[
    \mathbb{P}\left(
    \max_{i\ne j}|\langle \vv_i,\vv_j\rangle|>\frac{1}{4k}
    \right)
    \le
    \frac{3m^2}{2}\exp\left(-\frac{a_0n}{16k^2}\right).
\]
Choosing $C$ sufficiently large makes this failure probability strictly
smaller than one, so with positive probability the realization satisfies
\[
    |\langle \vv_i,\vv_j\rangle|
    \le
    \frac{1}{4k}
    \qquad\text{for all }i\ne j.
    \tag{$\star$}
\]
Positive probability is enough for an existential upper bound. For a
fixed finite instance, increasing $C$ makes the displayed union-bound
failure probability arbitrarily small.

Fix a realization satisfying $(\star)$. For any nonempty
$S\subseteq[m]$ with $s:=|S|\le k$, use the unnormalized centroid and
its normalized direction
\[
    \vz_S=\sum_{\ell\in S}\vv_\ell,
    \qquad
    \vu_S=\frac{\vz_S}{\|\vz_S\|_2}.
\]
Dividing $\vz_S$ by $s$ gives the arithmetic centroid but does not
change the normalized query direction.

For a selected object $i\in S$, the self term contributes one and the
remaining terms are controlled by $(\star)$:
\[
    \langle \vz_S,\vv_i\rangle
    =1+\sum_{\ell\in S,\ell\ne i}\langle \vv_\ell,\vv_i\rangle
    \ge
    1-\frac{s-1}{4k}
    \ge
    \frac{3}{4}.
\]
For an unselected object $j\notin S$, all terms are coherence noise:
\[
    \langle \vz_S,\vv_j\rangle
    =\sum_{\ell\in S}\langle \vv_\ell,\vv_j\rangle
    \le
    \frac{s}{4k}
    \le
    \frac{1}{4}.
\]
Thus every selected object beats every unselected object by at least
$1/2$ before normalizing the query.

It remains to bound the normalization factor. Again using $(\star)$,
\[
\begin{aligned}
    \|\vz_S\|_2^2
    &=
    s+2\sum_{\ell<r,\,\ell,r\in S}
      \langle \vv_\ell,\vv_r\rangle \\
    &\le
    s+\frac{s(s-1)}{4k}
    \le
    2k.
\end{aligned}
\]
Therefore, for every $i\in S$ and $j\notin S$,
\[
    \langle \vu_S,\vv_i\rangle-
    \langle \vu_S,\vv_j\rangle
    =
    \frac{\langle \vz_S,\vv_i\rangle-
          \langle \vz_S,\vv_j\rangle}{\|\vz_S\|_2}
    \ge
    \frac{1}{2\sqrt{2k}}.
\]
The same realization robustly retrieves every nonempty subset of size at
most $k$ using normalized centroid queries. Hence
\[
    \rmedc(m,k,c/\sqrt{k};\fln)
    \le
    n
    =O(k^2\log m)
\]
for a universal constant $c>0$, which proves
Theorem~\ref{thm:rmed-gaussian}.
\end{proof}

\subsection{Proof of Proposition~\ref{prop:rmedc-consequences-main}}\label{app:proof:prop:rmedc-consequences-main}

\begin{proof}
The linear-score statement is exactly Theorem~\ref{thm:rmed-gaussian}.
For cosine similarity, all objects and centroid query directions are
unit vectors, so cosine scores equal inner products and the same margin
is preserved. For Euclidean distance, let $i\in S$ and $j\notin S$.
The squared-distance identity gives
\[
    \|\vv_j-\vu_S\|_2^2-\|\vv_i-\vu_S\|_2^2
    =
    2\langle \vu_S,\vv_i-\vv_j\rangle.
\]
Thus an inner-product margin $\epsilon$ gives a squared-distance gap
$2\epsilon$. Since both distances are between unit vectors, their sum is
at most $4$, and the ordinary distance gap is at least $\epsilon/2$.
This gives the Euclidean robust centroid-query statement after changing
the universal constant.
\end{proof}

\subsection{Exact centroid-query consequences}\label{app:medc-consequences}

The following exact-only MED-C consequences are implied by the robust centroid witness in Theorem~\ref{thm:rmed-gaussian}. We keep the direct proof here because it records the simple zero-margin calculation for the three score functions.

\begin{theorem}\label{thm:medc-linear}
    $\medc(m, k; s_{\rm linear})\leq O(k^2\log m).$
\end{theorem}
\begin{theorem}\label{thm:medc-cos}
    $\medc(m, k; s_{\cos})\leq O(k^2\log m).$
\end{theorem}
\begin{theorem}\label{thm:medc-l2}
    $\medc(m, k; s_{\ell_2})\leq O(k^2\log m).$
\end{theorem}

\begin{proof}[Proof of Theorems~\ref{thm:medc-linear}--\ref{thm:medc-l2}]
We prove the three statements by constructing one set of object embeddings that works for all three scoring functions. Let
\[
    n = \left\lceil C k^2\log m\right\rceil
\]
for a sufficiently large universal constant $C$. Sample independent Gaussian vectors
\[
    g_1,\dots,g_m\sim \mathcal{N}(0,I_n/n),
    \qquad
    \vv_i=\frac{g_i}{\|g_i\|_2}.
\]
Then $\vv_1,\dots,\vv_m$ are independent unit vectors. For any fixed pair $i\ne j$, standard concentration for random unit vectors gives
\[
    \mathbb{P}\!\left(
    |\langle \vv_i,\vv_j\rangle|>\frac{1}{4k}
    \right)
    \leq
    2\exp\!\left(-c_0 n/k^2\right)
\]
for a universal constant $c_0>0$. A union bound over all pairs shows that, for $C$ large enough, the following coherence event has positive probability:
\[
    |\langle \vv_i,\vv_j\rangle|
    \leq
    \frac{1}{4k}
    \qquad \forall i\ne j.
    \tag{$\star$}
\]
Positive probability suffices for the existential MED-C upper bound. For any fixed finite instance, increasing $C$ makes the union-bound failure probability as small as desired.
Fix a realization satisfying $(\star)$.

Let $S\subseteq X$ be any nonempty subset with $s:=|S|\leq k$, and define its unnormalized and normalized centroids
\[
    \vz_S=\sum_{\ell\in S}\vv_\ell,
    \qquad
    \vc_S=\frac{1}{s}\vz_S .
\]
For a selected object $i\in S$,
\[
    \langle \vz_S,\vv_i\rangle
    =
    1+\sum_{\ell\in S,\ell\ne i}\langle \vv_\ell,\vv_i\rangle
    \geq
    1-\frac{s-1}{4k}
    \geq
    \frac{3}{4}.
\]
For an unselected object $j\notin S$,
\[
    \langle \vz_S,\vv_j\rangle
    =
    \sum_{\ell\in S}\langle \vv_\ell,\vv_j\rangle
    \leq
    \frac{s}{4k}
    \leq
    \frac{1}{4}.
\]
Therefore
\begin{align}
    \langle \vz_S,\vv_i\rangle
    -
    \langle \vz_S,\vv_j\rangle
    &\geq
    \frac{1}{2}, \notag\\
    \langle \vc_S,\vv_i\rangle
    -
    \langle \vc_S,\vv_j\rangle
    &\geq
    \frac{1}{2s}>0.
    \label{eq:medc-centroid-gap}
\end{align}
Thus the centroid $\vc_S$ ranks every selected object above every unselected object by inner product, proving Theorem~\ref{thm:medc-linear}.

For cosine similarity, the same centroid query is nonzero. Indeed, $(\star)$ gives
\[
    \|\vz_S\|_2^2
    =
    s+2\sum_{\ell<r,\ \ell,r\in S}\langle \vv_\ell,\vv_r\rangle
    \geq
    s-\frac{s(s-1)}{4k}
    \geq
    \frac{3s}{4}.
\]
Because each object vector has unit norm, for fixed $S$,
\[
    s_{\cos}(\vv_i,\vc_S)
    =
    \frac{\langle \vv_i,\vc_S\rangle}{\|\vc_S\|_2}
\]
differs from the linear score only by the positive factor $\|\vc_S\|_2^{-1}$. Hence the strict ordering in \eqref{eq:medc-centroid-gap} is preserved, proving Theorem~\ref{thm:medc-cos}.

For Euclidean distance, compare squared distances to the same centroid. Since all object vectors have norm one,
\[
    \|\vv_i-\vc_S\|_2^2
    =
    1+\|\vc_S\|_2^2-2\langle \vv_i,\vc_S\rangle .
\]
The terms $1+\|\vc_S\|_2^2$ are identical for all objects in the query $S$. Therefore \eqref{eq:medc-centroid-gap} implies
\[
    \|\vv_i-\vc_S\|_2^2
    <
    \|\vv_j-\vc_S\|_2^2
    \qquad
    \forall i\in S,\ j\notin S,
\]
or equivalently $s_{\ell_2}(\vv_i,\vc_S)>s_{\ell_2}(\vv_j,\vc_S)$. This proves Theorem~\ref{thm:medc-l2}.

We have exhibited, with positive probability, a configuration in dimension $n=O(k^2\log m)$ that satisfies the centroid-query MED condition under all three scoring functions. Hence the three MED-C upper bounds follow.
\end{proof}
\section{Additional Theoretical Comparisons}\label{app:additional-comparisons}

The common thread in the comparisons below is that exact realizability, positive-margin realizability, and learned realizability are different questions. The main body focuses on the exact MED and RMED statements; this appendix records the longer comparisons that are useful for interpreting nearby theory and empirical evidence.

\subsection{Sign-Rank, Exact Separability, and Positive Margins}

\citet{weller2025theoretical} also discuss sign-rank formulations for embedding-based retrieval. This is the right language for exact binary relevance matrices. Let $A\in\{0,1\}^{Q\times m}$ be a query-object relevance matrix and let $M=2A-1\in\{-1,+1\}^{Q\times m}$ be its sign matrix. If object vectors $\vx_i\in\real^d$, query vectors $\vw_q\in\real^d$, and thresholds $b_q$ exactly realize $A$ by
\[
    A_{q i}=1
    \quad\Longleftrightarrow\quad
    \langle \vw_q,\vx_i\rangle>b_q,
\]
then the augmented vectors $(\vw_q,-b_q)$ and $(\vx_i,1)$ give a sign factorization in dimension $d+1$. Thus the usual sign-rank of $M$ is at most $d+1$. Conversely, a rank-$r$ sign factorization of $M$ gives homogeneous inner-product thresholding in dimension $r$, up to this affine/homogeneous one-dimension convention.

This sign-rank viewpoint is an exact-separability characterization. It can certify or obstruct the existence of some threshold realization of a fixed binary relevance matrix, but it does not impose a normalized positive margin. For the full at-most-$k$ subset-incidence matrix whose rows are indexed by $S\in\collection_k$, our cyclic-polytopal construction gives an explicit exact realization in $2k$ dimensions for inner-product scoring, so the corresponding augmented sign-rank is at most $2k+1$. This exact witness is fully compatible with Weller et al.'s positive-margin lower bound: RMED asks whether the same kind of relevance family can be realized with normalized objects, normalized queries, and a uniform score gap. That stronger margin requirement is what makes $m$ reappear.

\subsection{What the Empirical Evidence Does and Does Not Debunk}

The experiments in Section~\ref{sec:maed_and_exp} debunk a specific empirical interpretation: observed failures of learned or free-optimization protocols should not be read as evidence that exact low-dimensional geometric witnesses do not exist. They do not debunk the positive-margin lower bound of \citet{weller2025theoretical}, nor do they imply that practical learned retrieval is easy.

For the synthetic top-$2$ setting, the cyclic-polytope construction gives exact dimension-$4$ witnesses for arbitrary top-$2$ answer sets. Therefore, a failed free-embedding optimization run is not evidence of exact geometric impossibility. It is evidence about the optimization protocol, query learning, numerical conditioning, margins, or finite precision.

For LIMIT and LIMIT-small, the packaged top-$2$ instances also have exact dimension-$4$ cyclic-polytope witnesses. In addition, the label-unaware random token-sum controls exceed the strongest comparable single-vector Promptriever baseline reported by \citet{weller2025theoretical} on Recall@2. These controls are not deployable retrieval models, but they show that the benchmark admits strong single-vector solutions under simple constructions. Thus the hard-dataset evidence does not establish lack of exact single-vector geometric capacity.

In this limited sense, the experiments rule out exact geometric approximability as the explanation for those failures. The remaining explanations are the stronger requirements emphasized throughout the paper: robustness, conditioning, tokenization, query-map learnability, objective choice, and optimization protocol.

\subsection{Structured Retrieval Settings: Multiclass and Hierarchical Retrieval}

\citet{guo2019breaking} study multiclass embedding classification. A naive translation of arbitrary at-most-$k$ answer sets into their language would create
\[
    N_k=\sum_{s=1}^{k}\binom{m}{s}
\]
possible answer-set labels, not merely $\binom{m}{k}$ unless the problem is restricted to exactly-$k$ answers. The number of answer-set labels containing any fixed object would be
\[
    R_k=\sum_{s=1}^{k}\binom{m-1}{s-1}.
\]
Plugging $N_k$ and $R_k$ into a multiclass bound is only a structural comparison, not a theorem for MED. Their setting controls a different classification protocol, whereas MED asks whether the complete arbitrary at-most-$k$ set family can be threshold-realized by query-specific functionals. The cyclic-polytopal construction answers that exact MED question directly.

\citet{you2025hierarchical} study hierarchical retrieval, where answer sets are reachable sets in a directed acyclic graph. This is also structurally related but different. The arbitrary at-most-$k$ family studied by MED has $\sum_{s=1}^{k}\binom{m}{s}$ possible answer sets; for fixed $k$, the exactly-$k$ term satisfies $\binom{m}{k}=\Theta(m^k/k!)$. Hierarchical retrieval instead restricts answer sets to graph-reachable families tied to source objects and the graph structure. Those reachable families are much smaller and more structured than all arbitrary subsets up to size $k$. Consequently, HR bounds should not be read as direct upper or lower bounds for arbitrary top-$k$ MED.

These structured settings are valuable comparisons because they also relate dimension to retrieval families. They should not be read as direct bounds for MED, whose object is the complete family of arbitrary answer sets up to size $k$.
\section{Random Token-Sum LIMIT Retrieval}\label{app:random-token-limit}

The LIMIT retrieval curves in Figure~\ref{fig:limit-retrieval} use the
label-unaware random token-sum construction in
Algorithm~\ref{alg:random-token-limit}. The tokenizer $\tau$ is one of
the handmade phrase tokenizer, the compact observed Qwen-tokenizer
vocabulary, or the vanilla word tokenizer. In all cases, random vectors
are allocated only for tokens observed in the LIMIT documents and
queries; the full Qwen vocabulary is not materialized. Repeated tokens
are counted in the sum. The implementation streams document embeddings
in chunks on the full LIMIT setting when needed, which changes memory
use but not the score definition below. Seeds are recorded in the
released CSV artifact; all current rows use seed $42+d$ except the full
LIMIT vanilla row at $d=64$, which uses seed $107$.

\input{tables/limit_promptriever_crossing}

\begin{algorithm}[H]
\caption{Label-unaware random token-sum retrieval}
\label{alg:random-token-limit}
\begin{algorithmic}[1]
\REQUIRE Corpus $C$, queries $Q$, qrels $R$, tokenizer $\tau$, dimension $d$, seed $s$
\STATE Build observed token set $V=\bigcup_{x\in C\cup Q}\tau(x)$ and assign compact ids to $V$
\STATE Draw $G_t\in\mathbb{R}^d$ independently from $\mathcal{N}(0,I_d)$ for each $t\in V$, using seed $s$
\FORALL{$x\in C\cup Q$}
  \STATE Embed $x$ as $\phi(x)=\sum_{t\in\tau(x)}G_t$
\ENDFOR
\FORALL{$q\in Q$}
  \STATE Score each document $c\in C$ by $S(q,c)=\langle \phi(q),\phi(c)\rangle$
  \STATE Let $T_2(q)$ be the two highest-scoring documents
  \STATE Accumulate $\left|T_2(q)\cap R(q)\right|/\left|R(q)\right|$
\ENDFOR
\ENSURE Mean accumulated value over queries, reported as Recall@2
\end{algorithmic}
\end{algorithm}

\begin{table}[H]
  \centering
  \small
  \caption{Full random token-sum LIMIT Recall@2 grid used in
  Figure~\ref{fig:limit-retrieval}.}
  \label{tab:limit-random-token-grid}
  \input{tables/limit_retrieval_table}
\end{table}

\clearpage

\section{Cyclic-Polytope LIMIT Embedding Tables}\label{app:cyclic-limit}

The cyclic-polytope construction embeds the LIMIT documents on the
moment curve in $\mathbb{R}^4$ and constructs each query with the
squared polynomial that vanishes on its two relevant document
parameters. Table~\ref{tab:limit-cyclic-overfit-summary} reports the
exact overfit summary for the packaged LIMIT and LIMIT-small assets in
dimension $4$. For readability, the exported LIMIT-small document
parameters are linearly spaced in $[-1,1]$, so the listed document
vectors lie in $[-1,1]^4$; each listed query vector is rescaled by a
positive scalar to also lie in $[-1,1]^4$, which does not affect top-$2$
rankings.

\begin{table}[H]
  \centering
  \caption{Cyclic-polytope overfit summary for packaged LIMIT assets.}
  \label{tab:limit-cyclic-overfit-summary}
  \input{tables/limit_cyclic_overfit_summary}
\end{table}
Tables~\ref{tab:limit-small-cyclic-doc-embeddings}
and~\ref{tab:limit-small-cyclic-query-embeddings} list the resulting
document embeddings and 50 selected query embeddings. The numeric
\emph{doc id} is the index used in the construction; the profile id is
the original LIMIT corpus identifier, which is a person/profile name.

\clearpage
\input{tables/limit_small_cyclic_document_embeddings_table}

\clearpage
\input{tables/limit_small_cyclic_query_embeddings_table}

\clearpage
\end{document}

%% file: custom_command.tex
\usepackage{microtype}
\usepackage{graphicx}
\usepackage{subcaption}
\usepackage{booktabs}
\usepackage{amsmath,amsfonts,amssymb,amsthm,bm}
\usepackage{paralist}
\usepackage{xcolor}
\usepackage{tcolorbox}
\usepackage{multirow}
\usepackage{longtable}
\usepackage{hyperref}

\usepackage[accepted]{icml2026}

\theoremstyle{plain}
\newtheorem{theorem}{Theorem}[section]
\newtheorem{proposition}[theorem]{Proposition}
\newtheorem{lemma}[theorem]{Lemma}
\theoremstyle{definition}
\newtheorem{definition}[theorem]{Definition}
\newtheorem{example}[theorem]{Example}
\theoremstyle{remark}
\newtheorem{remark}[theorem]{Remark}

\newcommand{\collection}{\mathcal{C}}
\newcommand{\real}{\mathbb{R}}

\newcommand{\med}{\textsc{MED}}
\newcommand{\medc}{\textsc{MED-C}}
\newcommand{\rmed}{\textsc{RMED}}
\newcommand{\rmedc}{\textsc{RMED-C}}
\newcommand{\vcd}{\textsc{VCD}}
\newcommand{\functionals}{\mathcal{F}}
\newcommand{\fln}{\functionals_{\rm linear}}
\newcommand{\fcos}{\functionals_{\cos}}

\newcommand{\fl}[1]{\functionals_{\ell_{#1}}}

\newcommand{\vc}{\bm{c}}
\newcommand{\vq}{\bm{q}}
\newcommand{\vu}{\bm{u}}
\newcommand{\vv}{\bm{v}}
\newcommand{\vw}{\bm{w}}
\newcommand{\vx}{\bm{x}}
\newcommand{\vy}{\bm{y}}
\newcommand{\vz}{\bm{z}}

%% file: tables/limit_promptriever_crossing.tex
\begin{table}[H]
  \centering
  \small
  \caption{Random token-sum controls compared with Weller et al.'s
  strongest comparable single-vector baseline, Promptriever Llama3 8B
  at $d=4096$ (we use baseline in the table).}
  \label{tab:limit-promptriever-crossing}
  \begin{tabular}{lrlrr}
    \toprule
    Dataset & Baseline R@2 & Tokenizer & First $d$ above baseline & R@2 at $d=4096$ \\
    \midrule
    LIMIT & 0.030 & handmade & 256 & 0.9980 \\
    LIMIT & 0.030 & qwen & 512 & 0.2675 \\
    LIMIT & 0.030 & vanilla & 512 & 0.7060 \\
    LIMIT-small & 0.543 & handmade & 256 & 1.0000 \\
    LIMIT-small & 0.543 & qwen & 1024 & 0.8010 \\
    LIMIT-small & 0.543 & vanilla & 512 & 0.9545 \\
    \bottomrule
  \end{tabular}
\end{table}

%% file: tables/limit_retrieval_table.tex
\begin{tabular}{llrr}
\toprule
Dataset & Tokenizer & $d$ & Recall@2 \\
\midrule
LIMIT & handmade & 32 & 0.0020 \\
LIMIT & handmade & 64 & 0.0045 \\
LIMIT & handmade & 128 & 0.0140 \\
LIMIT & handmade & 256 & 0.0725 \\
LIMIT & handmade & 512 & 0.3035 \\
LIMIT & handmade & 1024 & 0.7725 \\
LIMIT & handmade & 2048 & 0.9915 \\
LIMIT & handmade & 4096 & 0.9980 \\
LIMIT & qwen & 32 & 0.0015 \\
LIMIT & qwen & 64 & 0.0010 \\
LIMIT & qwen & 128 & 0.0025 \\
LIMIT & qwen & 256 & 0.0120 \\
LIMIT & qwen & 512 & 0.0365 \\
LIMIT & qwen & 1024 & 0.0570 \\
LIMIT & qwen & 2048 & 0.1515 \\
LIMIT & qwen & 4096 & 0.2675 \\
LIMIT & vanilla & 32 & 0.0025 \\
LIMIT & vanilla & 64 & 0.0025 \\
LIMIT & vanilla & 128 & 0.0095 \\
LIMIT & vanilla & 256 & 0.0265 \\
LIMIT & vanilla & 512 & 0.0630 \\
LIMIT & vanilla & 1024 & 0.1990 \\
LIMIT & vanilla & 2048 & 0.4570 \\
LIMIT & vanilla & 4096 & 0.7060 \\
LIMIT-small & handmade & 32 & 0.1715 \\
LIMIT-small & handmade & 64 & 0.2670 \\
LIMIT-small & handmade & 128 & 0.4005 \\
LIMIT-small & handmade & 256 & 0.6215 \\
LIMIT-small & handmade & 512 & 0.8780 \\
LIMIT-small & handmade & 1024 & 0.9940 \\
LIMIT-small & handmade & 2048 & 1.0000 \\
LIMIT-small & handmade & 4096 & 1.0000 \\
LIMIT-small & qwen & 32 & 0.1120 \\
LIMIT-small & qwen & 64 & 0.1695 \\
LIMIT-small & qwen & 128 & 0.2180 \\
LIMIT-small & qwen & 256 & 0.3735 \\
LIMIT-small & qwen & 512 & 0.5160 \\
LIMIT-small & qwen & 1024 & 0.6010 \\
LIMIT-small & qwen & 2048 & 0.7510 \\
LIMIT-small & qwen & 4096 & 0.8010 \\
LIMIT-small & vanilla & 32 & 0.1230 \\
LIMIT-small & vanilla & 64 & 0.1705 \\
LIMIT-small & vanilla & 128 & 0.2925 \\
LIMIT-small & vanilla & 256 & 0.3845 \\
LIMIT-small & vanilla & 512 & 0.6145 \\
LIMIT-small & vanilla & 1024 & 0.7845 \\
LIMIT-small & vanilla & 2048 & 0.8980 \\
LIMIT-small & vanilla & 4096 & 0.9545 \\
\bottomrule
\end{tabular}

%% file: tables/limit_cyclic_overfit_summary.tex
\begin{tabular}{lrrrrr}
\toprule
Dataset & Docs & Queries & Qrels & Minimal $d$ & Recall@2 \\
\midrule
LIMIT-small & 46 & 1000 & 2000 & 4 & 1 \\
LIMIT & 50000 & 1000 & 2000 & 4 & 1 \\
\bottomrule
\end{tabular}

%% file: tables/limit_small_cyclic_document_embeddings_table.tex
{\scriptsize
\setlength{\tabcolsep}{3pt}
\begin{longtable}{@{}r l r r r r@{}}
\caption{LIMIT-small cyclic-polytope document embeddings in $[-1,1]^4$.}\label{tab:limit-small-cyclic-doc-embeddings}\\
\toprule
Doc id & Profile id & $x_1$ & $x_2$ & $x_3$ & $x_4$ \\
\midrule
\endfirsthead
\caption[]{LIMIT-small cyclic-polytope document embeddings (continued).}\\
\toprule
Doc id & Profile id & $x_1$ & $x_2$ & $x_3$ & $x_4$ \\
\midrule
\endhead
0 & Geneva Durben & -1 & 1 & -1 & 1 \\
1 & Dorathea Bastress & -0.9556 & 0.9131 & -0.8725 & 0.8337 \\
2 & Armand Schweda & -0.9111 & 0.8301 & -0.7563 & 0.6891 \\
3 & Flor Lemaire & -0.8667 & 0.7511 & -0.6510 & 0.5642 \\
4 & Pate Lindley & -0.8222 & 0.6760 & -0.5559 & 0.4570 \\
5 & Shelvia Goike & -0.7778 & 0.6049 & -0.4705 & 0.3660 \\
6 & Ovid Rahm & -0.7333 & 0.5378 & -0.3944 & 0.2892 \\
7 & Bronson Saelee & -0.6889 & 0.4746 & -0.3269 & 0.2252 \\
8 & Gladstone Oonk & -0.6444 & 0.4153 & -0.2676 & 0.1725 \\
9 & Ofelia Rosselot & -0.6000 & 0.3600 & -0.2160 & 0.1296 \\
10 & Tisha Ghent & -0.5556 & 0.3086 & -0.1715 & 0.0953 \\
11 & Herminia Caranto & -0.5111 & 0.2612 & -0.1335 & 0.0682 \\
12 & Linzy Recknor & -0.4667 & 0.2178 & -0.1016 & 0.0474 \\
13 & Vinie Relford & -0.4222 & 0.1783 & -0.0753 & 0.0318 \\
14 & Jerrod Dumpit & -0.3778 & 0.1427 & -0.0539 & 0.0204 \\
15 & Amaris Grow & -0.3333 & 0.1111 & -0.0370 & 0.0123 \\
16 & Marcellus Meachum & -0.2889 & 0.0835 & -0.0241 & 0.0070 \\
17 & Wellington Hinn & -0.2444 & 0.0598 & -0.0146 & 0.0036 \\
18 & Georgette Cagna & -0.2000 & 0.0400 & -0.0080 & 0.0016 \\
19 & Laurine Bellizzi & -0.1556 & 0.0242 & -0.0038 & 0.0006 \\
20 & Agnes Reap & -0.1111 & 0.0123 & -0.0014 & 0.0002 \\
21 & Sheree Riddley & -0.0667 & 0.0044 & -0.0003 & 0.0000 \\
22 & Mathew Weierke & -0.0222 & 0.0005 & -0.0000 & 0.0000 \\
23 & Casimiro Steo & 0.0222 & 0.0005 & 0.0000 & 0.0000 \\
24 & Maryann Bohnsack & 0.0667 & 0.0044 & 0.0003 & 0.0000 \\
25 & Flo Zaugg & 0.1111 & 0.0123 & 0.0014 & 0.0002 \\
26 & Nathen Saadia & 0.1556 & 0.0242 & 0.0038 & 0.0006 \\
27 & Ruby Gaskin & 0.2000 & 0.0400 & 0.0080 & 0.0016 \\
28 & Jerrie Roupe & 0.2444 & 0.0598 & 0.0146 & 0.0036 \\
29 & Camisha Bogosian & 0.2889 & 0.0835 & 0.0241 & 0.0070 \\
30 & Gaetano Argel & 0.3333 & 0.1111 & 0.0370 & 0.0123 \\
31 & Nathaniel Robens & 0.3778 & 0.1427 & 0.0539 & 0.0204 \\
32 & Tarik Hollfelder & 0.4222 & 0.1783 & 0.0753 & 0.0318 \\
33 & Riya Hayhoe & 0.4667 & 0.2178 & 0.1016 & 0.0474 \\
34 & Chaney Gertman & 0.5111 & 0.2612 & 0.1335 & 0.0682 \\
35 & Cristy Walford & 0.5556 & 0.3086 & 0.1715 & 0.0953 \\
36 & Eustace Comment & 0.6000 & 0.3600 & 0.2160 & 0.1296 \\
37 & Terrell Varadarajan & 0.6444 & 0.4153 & 0.2676 & 0.1725 \\
38 & Darwyn Raio & 0.6889 & 0.4746 & 0.3269 & 0.2252 \\
39 & Eudora Cervero & 0.7333 & 0.5378 & 0.3944 & 0.2892 \\
40 & Jacey Gnatek & 0.7778 & 0.6049 & 0.4705 & 0.3660 \\
41 & Elam Mejiamejia & 0.8222 & 0.6760 & 0.5559 & 0.4570 \\
42 & Celia Marszalek & 0.8667 & 0.7511 & 0.6510 & 0.5642 \\
43 & Aliza Uhlrich & 0.9111 & 0.8301 & 0.7563 & 0.6891 \\
44 & Chadwick Frisella & 0.9556 & 0.9131 & 0.8725 & 0.8337 \\
45 & Theola Laudermilk & 1 & 1 & 1 & 1 \\
\bottomrule
\end{longtable}
}

%% file: tables/limit_small_cyclic_query_embeddings_table.tex
{\scriptsize
\setlength{\tabcolsep}{3pt}
\begin{longtable}{@{}r l l r r r r@{}}
\caption{Selected LIMIT-small squared-polynomial query embeddings normalized into $[-1,1]^4$.}\label{tab:limit-small-cyclic-query-embeddings}\\
\toprule
Query & Query id & Positive doc ids & $q_1$ & $q_2$ & $q_3$ & $q_4$ \\
\midrule
\endfirsthead
\caption[]{Selected LIMIT-small squared-polynomial query embeddings (continued).}\\
\toprule
Query & Query id & Positive doc ids & $q_1$ & $q_2$ & $q_3$ & $q_4$ \\
\midrule
\endhead
0 & query\_0 & 0, 1 & -0.6516 & -1 & -0.6819 & -0.1744 \\
20 & query\_20 & 0, 21 & -0.0667 & -0.5958 & -1 & -0.4688 \\
40 & query\_40 & 0, 41 & 0.1813 & 1 & -0.2205 & -0.6200 \\
61 & query\_61 & 1, 18 & -0.1911 & -0.7432 & -1 & -0.4327 \\
81 & query\_81 & 1, 38 & 0.2819 & 1 & -0.4282 & -0.8029 \\
101 & query\_101 & 2, 15 & -0.3037 & -0.8663 & -1 & -0.4018 \\
122 & query\_122 & 2, 36 & 0.3401 & 0.9965 & -0.6222 & -1 \\
142 & query\_142 & 3, 14 & -0.3274 & -0.8853 & -1 & -0.4018 \\
163 & query\_163 & 3, 35 & 0.2996 & 0.8662 & -0.6222 & -1 \\
183 & query\_183 & 4, 14 & -0.3106 & -0.8588 & -1 & -0.4167 \\
203 & query\_203 & 4, 34 & 0.2615 & 0.7437 & -0.6222 & -1 \\
224 & query\_224 & 5, 15 & -0.2593 & -0.7889 & -1 & -0.4500 \\
244 & query\_244 & 5, 35 & 0.1920 & 0.8148 & -0.4444 & -1 \\
265 & query\_265 & 6, 17 & -0.1793 & -0.6722 & -1 & -0.5114 \\
285 & query\_285 & 6, 37 & 0.0840 & 0.9373 & -0.1778 & -1 \\
305 & query\_305 & 7, 19 & -0.1072 & -0.5491 & -1 & -0.5921 \\
326 & query\_326 & 7, 40 & -0.0895 & 1 & 0.1671 & -0.9401 \\
346 & query\_346 & 8, 23 & 0.0143 & -0.2881 & -1 & -0.8036 \\
366 & query\_366 & 8, 43 & -0.2839 & 1 & 0.4834 & -0.9064 \\
387 & query\_387 & 9, 28 & 0.1043 & 0.1669 & -0.7111 & -1 \\
407 & query\_407 & 10, 13 & -0.2346 & -0.7288 & -1 & -0.5114 \\
428 & query\_428 & 10, 34 & 0.0252 & 0.5659 & -0.0889 & -1 \\
448 & query\_448 & 11, 20 & -0.0568 & -0.4024 & -1 & -0.8036 \\
468 & query\_468 & 11, 40 & -0.2120 & 0.7240 & 0.5333 & -1 \\
489 & query\_489 & 12, 28 & 0.0507 & 0.1788 & -0.4444 & -1 \\
509 & query\_509 & 13, 16 & -0.1220 & -0.5271 & -1 & -0.7031 \\
530 & query\_530 & 13, 37 & -0.1209 & 0.4948 & 0.4444 & -1 \\
550 & query\_550 & 14, 26 & 0.0261 & 0.0681 & -0.4444 & -1 \\
570 & query\_570 & 15, 16 & -0.0963 & -0.4659 & -1 & -0.8036 \\
591 & query\_591 & 15, 37 & -0.1337 & 0.3328 & 0.6222 & -1 \\
611 & query\_611 & 16, 28 & 0.0063 & 0.1393 & -0.0889 & -1 \\
632 & query\_632 & 17, 21 & -0.0101 & -0.1294 & -0.6222 & -1 \\
652 & query\_652 & 17, 41 & -0.2010 & 0.0590 & 1 & -0.8654 \\
672 & query\_672 & 18, 34 & -0.0636 & 0.1077 & 0.6222 & -1 \\
693 & query\_693 & 19, 29 & -0.0120 & 0.0721 & 0.2667 & -1 \\
713 & query\_713 & 20, 24 & 0.0007 & 0.0128 & -0.0889 & -1 \\
733 & query\_733 & 20, 44 & -0.1062 & -0.2965 & 1 & -0.5921 \\
754 & query\_754 & 21, 41 & -0.0548 & -0.3052 & 1 & -0.6618 \\
774 & query\_774 & 22, 38 & -0.0153 & -0.3104 & 1 & -0.7500 \\
795 & query\_795 & 23, 37 & 0.0143 & -0.3548 & 1 & -0.7500 \\
815 & query\_815 & 24, 36 & 0.0400 & -0.3933 & 1 & -0.7500 \\
835 & query\_835 & 25, 36 & 0.0667 & -0.4493 & 1 & -0.7031 \\
856 & query\_856 & 26, 38 & 0.1072 & -0.5491 & 1 & -0.5921 \\
876 & query\_876 & 27, 40 & 0.1556 & -0.6480 & 1 & -0.5114 \\
897 & query\_897 & 28, 44 & 0.2336 & -0.7947 & 1 & -0.4167 \\
917 & query\_917 & 30, 33 & 0.1556 & -0.5944 & 1 & -0.6250 \\
937 & query\_937 & 31, 39 & 0.2770 & -0.8049 & 1 & -0.4500 \\
958 & query\_958 & 33, 35 & 0.2593 & -0.7647 & 1 & -0.4891 \\
978 & query\_978 & 34, 44 & 0.4580 & -1 & 0.9378 & -0.3197 \\
999 & query\_999 & 37, 38 & 0.4440 & -0.9996 & 1 & -0.3750 \\
\bottomrule
\end{longtable}
}

%% file: preprint.bib
@inproceedings{weller2025theoretical,
  title={On the Theoretical Limitations of Embedding-Based Retrieval},
  author={Weller, Orion and Boratko, Michael and Naim, Iftekhar and Lee, Jinhyuk},
  booktitle={The Thirteenth International Conference on Learning Representations},
  year={2026},
  url={https://openreview.net/forum?id=k9CzIvzfaA},
  eprint={2508.21038},
  archivePrefix={arXiv},
  primaryClass={cs.IR}
}

@book{vershynin2018high,
  title={High-dimensional probability: An introduction with applications in data science},
  author={Vershynin, Roman},
  volume={47},
  year={2018},
  publisher={Cambridge university press}
}

@book{ziegler2012lectures,
  title={Lectures on polytopes},
  author={Ziegler, G{\"u}nter M},
  volume={152},
  year={2012},
  publisher={Springer Science \& Business Media}
}

@book{mohri2018foundations,
  title={Foundations of machine learning},
  author={Mohri, Mehryar and Rostamizadeh, Afshin and Talwalkar, Ameet},
  year={2018},
  publisher={MIT press}
}

@book{matousek2013lectures,
  title={Lectures on discrete geometry},
  author={Matousek, Jiri},
  volume={212},
  year={2013},
  publisher={Springer Science \& Business Media}
}

@article{kingma2014adam,
  title={Adam: A method for stochastic optimization},
  author={Kingma, Diederik P},
  journal={arXiv preprint arXiv:1412.6980},
  year={2014}
}

@inproceedings{weller2025mfollowir,
  title={{mFollowIR: A Multilingual Benchmark for Instruction Following in Retrieval}},
  author={Weller, Orion and Chang, Benjamin and Yang, Eugene and Yarmohammadi, Mahsa and Barham, Samuel and MacAvaney, Sean and Cohan, Arman and Soldaini, Luca and Van Durme, Benjamin and Lawrie, Dawn J.},
  booktitle={Advances in Information Retrieval: 47th European Conference on Information Retrieval, ECIR 2025, Proceedings, Part II},
  series={Lecture Notes in Computer Science},
  volume={15573},
  pages={295--310},
  year={2025},
  publisher={Springer},
  doi={10.1007/978-3-031-88711-6_19}
}

@article{lee2019latent,
  title={Latent retrieval for weakly supervised open domain question answering},
  author={Lee, Kenton and Chang, Ming-Wei and Toutanova, Kristina},
  journal={arXiv preprint arXiv:1906.00300},
  year={2019}
}

@article{izacard2021unsupervised,
  title={Unsupervised dense information retrieval with contrastive learning},
  author={Izacard, Gautier and Caron, Mathilde and Hosseini, Lucas and Riedel, Sebastian and Bojanowski, Piotr and Joulin, Armand and Grave, Edouard},
  journal={arXiv preprint arXiv:2112.09118},
  year={2021}
}

@article{wang2022text,
  title={Text embeddings by weakly-supervised contrastive pre-training},
  author={Wang, Liang and Yang, Nan and Huang, Xiaolong and Jiao, Binxing and Yang, Linjun and Jiang, Daxin and Majumder, Rangan and Wei, Furu},
  journal={arXiv preprint arXiv:2212.03533},
  year={2022}
}

@article{andoni2008near,
  title={Near-optimal hashing algorithms for approximate nearest neighbor in high dimensions},
  author={Andoni, Alexandr and Indyk, Piotr},
  journal={Communications of the ACM},
  volume={51},
  number={1},
  pages={117--122},
  year={2008},
  publisher={ACM New York, NY, USA}
}

@inproceedings{reimers2021curse,
  title={The curse of dense low-dimensional information retrieval for large index sizes},
  author={Reimers, Nils and Gurevych, Iryna},
  booktitle={Proceedings of the 59th Annual Meeting of the Association for Computational Linguistics and the 11th International Joint Conference on Natural Language Processing (Volume 2: Short Papers)},
  pages={605--611},
  year={2021}
}

@article{yin2018dimensionality,
  title={On the dimensionality of word embedding},
  author={Yin, Zi and Shen, Yuanyuan},
  journal={Advances in neural information processing systems},
  volume={31},
  year={2018}
}

@article{zaheer2017deep,
  title={Deep sets},
  author={Zaheer, Manzil and Kottur, Satwik and Ravanbakhsh, Siamak and Poczos, Barnabas and Salakhutdinov, Russ R and Smola, Alexander J},
  journal={Advances in neural information processing systems},
  volume={30},
  year={2017}
}

@inproceedings{aksoylar2014information,
  title={Information-theoretic bounds for adaptive sparse recovery},
  author={Aksoylar, Cem and Saligrama, Venkatesh},
  booktitle={2014 IEEE International Symposium on Information Theory},
  pages={1311--1315},
  year={2014},
  organization={IEEE}
}

@article{you2025hierarchical,
  title={Hierarchical Retrieval: The Geometry and a Pretrain-Finetune Recipe},
  author={You, Chong and Jayaram, Rajesh and Suresh, Ananda Theertha and Nittka, Robin and Yu, Felix and Kumar, Sanjiv},
  journal={arXiv preprint arXiv:2509.16411},
  year={2025}
}

@article{guo2019breaking,
  title={Breaking the glass ceiling for embedding-based classifiers for large output spaces},
  author={Guo, Chuan and Mousavi, Ali and Wu, Xiang and Holtmann-Rice, Daniel N and Kale, Satyen and Reddi, Sashank and Kumar, Sanjiv},
  journal={Advances in Neural Information Processing Systems},
  volume={32},
  year={2019}
}

@inproceedings{alon1985geometrical,
  title={Geometrical realization of set systems and probabilistic communication complexity},
  author={Alon, Noga and Frankl, Peter and Rodl, Vojtech},
  booktitle={26th Annual Symposium on Foundations of Computer Science (sfcs 1985)},
  pages={277--280},
  year={1985},
  organization={IEEE}
}
